\documentclass{article} 
\usepackage{arxiv}  
\usepackage[utf8]{inputenc} 
\usepackage[T1]{fontenc}    
\usepackage{hyperref}       
\usepackage{url}            
\usepackage{booktabs}       
\usepackage{amsfonts}       
\usepackage{nicefrac}       
\usepackage{microtype}      
\usepackage{lipsum} 
\usepackage{graphicx} 
\usepackage{algorithm}
\usepackage{algorithmicx}
\usepackage{amsmath,amssymb} 
\usepackage{float}
\usepackage{natbib}
\usepackage{array}
\usepackage{titlesec} 
\usepackage{caption}
\usepackage{float}

\usepackage{sectsty}
\usepackage{graphicx}
\usepackage{authblk}
\usepackage{amsmath}
\usepackage{natbib}
\usepackage{booktabs}
\usepackage{amsfonts}
\usepackage{amssymb}
\usepackage{rotating}
\usepackage{lipsum}
\usepackage{capt-of}
\usepackage{float}
\usepackage{placeins}

\usepackage{tabularx}
\titlespacing*{\subsection}{0pt}{*0}{*0}
\titlespacing*{\subsubsection}{0pt}{*0}{*0}
\titlespacing*{\section}{0pt}{*0}{*0}
\DeclareMathOperator*{\argmin}{arg\,min}

\begin{document}

\title{Conformal Prediction in Dynamic Biological Systems}  
\author{Alberto Portela\\ 
Computational Biology Lab\\ 
MBG-CSIC, Spanish National Research Council\\ 
Pontevedra, Spain \\ 
\texttt{alberto.portela@csic.es} 
\and 
Julio R. Banga\\ 
Computational Biology Lab\\ 
MBG-CSIC, Spanish National Research Council\\ 
Pontevedra, Spain \\ 
\texttt{j.r.banga@csic.es} 
\and 
Marcos Matabuena\\ 
Department of Biostatistics\\ 
Harvard University\\ 
Cambridge, MA, USA \\ 
\texttt{mmatabuena@hsph.harvard.edu} 
}  
\maketitle

\begin{abstract}     
 Uncertainty quantification (UQ) is the process of systematically determining and characterizing the degree of confidence in computational model predictions. In the context of systems biology, especially with dynamic models, UQ is crucial because it addresses the challenges posed by nonlinearity and parameter sensitivity, allowing us to properly understand and extrapolate the behavior of complex biological systems. Here, we focus on dynamic models represented by deterministic nonlinear ordinary differential equations. Many current UQ approaches in this field rely on Bayesian statistical methods. While powerful, these methods often require strong prior specifications and make parametric assumptions that may not always hold in biological systems. Additionally, these methods face challenges in domains where sample sizes are limited, and statistical inference becomes constrained, with computational speed being a bottleneck in large models of biological systems. As an alternative, we propose the use of conformal inference methods, introducing two novel algorithms that, in some instances, offer non-asymptotic guarantees, enhancing robustness and scalability across various applications. We demonstrate the efficacy of our proposed algorithms through several scenarios, highlighting their advantages over traditional Bayesian approaches. The proposed methods show promising results for diverse biological data structures and scenarios, offering a general framework to quantify uncertainty for dynamic models of biological systems.The software for the methodology and the reproduction of the results is available at \href{https://zenodo.org/doi/10.5281/zenodo.13644869}{https://zenodo.org/doi/10.5281/zenodo.13644869}.\\
\textbf{Contact:} \href{mailto:mmatabuena@hsph.harvard.edu}{mmatabuena@hsph.harvard.edu}; \href{mailto:j.r.banga@csic.es}{j.r.banga@csic.es}\\
\end{abstract}


\section{Introduction}



In the field of systems biology, we utilize mechanistic dynamic models as tools to analyze, predict and understand the intricate behaviors of complex biological processes \citep{albridge-burke-lauffenburger-sorger:2006,ingalls2013mathematical,distefano2015dynamic}. These models are typically composed of sets of deterministic nonlinear ordinary differential equations, and are designed to provide a quantitative understanding of the dynamics which would be difficult to achieve through other means. In particular, this mechanistic approach offers several advantages over data-driven approaches \citep{Coveney2016,Baker2018}. First, it can generate more accurate predictions and can be applied to a broader range of situations. Second, it provides a deeper understanding of how the system works thanks to its mechanism-based nature, making it easier to interpret the reasons behind its behavior. Finally, it requires less data for training because it is based on established theories and principles that describe the underlying processes. Overall, mechanistic models can help in understanding the dynamics of biological systems, in predicting their behaviour under different conditions, in generating testable hypotheses, and in identifying knowledge gaps.

However, these benefits come with a trade-off. As the number of elements (species) and unknown variables in the system increases, the model becomes significantly more complex in terms of number of parameters and non-linear relationships. This complexity can make it difficult to interpret the model's results \citep{Prybutok2022-kz}, and damages its identifiability, i.e. the ability to uniquely determine the unknown parameters in a model from the available data. As the complexity increases with more species and unknown parameters, achieving full identifiability and observability becomes more difficult \citep{villaverde2014reverse,Massonis2022}. As a result, developing a reliable dynamic mechanistic model can be a demanding and error prone task, requiring considerable expertise and the use of comprehensive and systematic protocols \citep{liepe2014framework,Villaverde2022protocol,Linden2022,Simpson2023}. Additionally, highly detailed models can lead to more uncertain predictions \citep{Puy2022-yc,Babtie2017-rs}. The uncertainty in parameters and their identifiability impacts model predictions, thus ideally we should be able to characterize such impact in an interpretable manner, aiming to make useful predictions despite poor identifiability \citep{cedersund:2012,Simpson2023}. Therefore, quantifying this uncertainty and how it affects different system states, a process known as Uncertainty Quantification (UQ) \citep{Smith2013}, is an open and fundamental challenge \citep{geris2016uncertainty,Mitra2019,sharp2022parameter,Villaverde2023_comparisonUQ}.

UQ plays a key role in enhancing the reliability and interpretability of mechanistic dynamic models \citep{kaltenbach2009systems,Mikovi2010,kirk2015systems,cedersund2016prediction}. It helps in understanding the underlying uncertainties in the model parameters and predictions, thereby improving the model’s predictive power and its utility in decision-making processes \citep{National_Research_Council2012-wq,begoli2019need}. A lack of proper UQ can result in models that are too confident in their predictions, which can be misleading. 

Different approaches for uncertainty quantification and robustness analysis in the context of systems biology have been reviewed elsewhere \citep{kaltenbach2009systems,cedersund-roll:2009,vanlier-tiemann-hilbers-vanriel:2013,streif2016robustness}. Roughly speaking, we can distinguish between Bayesian and frequentist methods \citep{Bayarri2004,liepe2014framework,Eriksson2019,Linden2022,Villaverde2022protocol,Murphy2024}. 
Lately, the most predominant approach in the literature is to use Bayesian methods, which treat model parameters as random variables. Bayesian methods can perform well even with small sample sizes, especially when informative priors are used. Frequentist methods often require larger sample sizes to achieve reliable estimates. In practice, Bayesian approaches require parametric assumptions to define likelihood equations and the specification of a prior to derive an approximate posterior distribution of parameters for approximate and analytical methods \citep{Eriksson2019,Linden2022}. This process is crucial for model estimation and inference.

However, Bayesian approaches often demand significant computational resources. Moreover, in the particular case of systems of differential equations, it is also typical to encounter identifiability issues, which result in multimodal posterior distributions that are challenging to handle in practice. In fact, although non-identifiabilities poses challenges to both frequentist and Bayesian sampling approaches, the latter can be especially susceptible to convergence failures \citep{Bayarri2004,raue2013joining,hines2014determination,plank2024structured}. Prediction profile likelihood methods and variants \citep{Hinkley1979,kreutz2012likelihood,hass2016fast,Simpson2023,Murphy2024} provide a competitive alternative by combining a frequentist perspective with a maximum projection of the likelihood by solving a sequence of optimization problems. However, they can be computationally demanding when a large number of predictions must be assessed.

Although various uncertainty quantification approaches are utilized in the domain of systems biology, comparative assessments of the strengths and weaknesses of state-of-the-art methods remain scarce. \cite{Villaverde2023_comparisonUQ} recently presented a systematic comparison of four methods: Fisher information matrix (FIM), Bayesian sampling, prediction profile likelihood and ensemble modelling. The comparison was made considering case studies of increasing computational complexity. This assessment revealed an interplay between their applicability and statistical interpretability. The FIM method was not reliable, and the prediction profile likelihood did not scale up well, being very computationally demanding when a large number of predictions had to be assessed. An interesting trade-off between computational scalability and accuracy and statistical guarantees was found for the ensemble and Bayesian sampling approaches. The Bayesian method proved adequate for less complex scenarios; however, it faced scalability challenges and encountered convergence difficulties when applied to the more intricate problems. The ensemble approach presented better performance for large-scale models, but weaker theoretical justification.

Therefore there is a clear need of UQ methods with both good scalability and strong theoretical statistical properties. Recently, in the literature of statistics and machine learning research, the use of conformal prediction \citep{shafer2008tutorial} to quantify the uncertainty of model outputs has become increasingly popular as an alternative to Bayesian methods and other asymptotic approximations \citep{lugosi2024uncertainty}. One of the successful aspects of this methodology in practice is the  non-asymptotic guarantees that ensure the coverage of prediction regions is well-calibrated, at least from a global (marginal) perspective, as reviewed by \cite{angelopoulos2023conformal}. However, to the best of our knowledge, their use in the systems biology and dynamical systems literature is not widespread, despite their expected promising properties for examining and making predictions in complex biological systems.

Considering systems biology applications, we must account for the typically limited number of observations. In order to increase the statistical modelling efficiency, we focus on conformal predictions based on the estimation of the conditional mean regression function \citep{lei2018distribution}, and the semi-parametric distributional location-scale regression models \citep{siegfried2023distribution}. To exploit the specific structure of location-scale regression models and maximize the limited available information, we propose two algorithms based on the jackknife methodology (see for example \cite{10.1214/20-AOS1965}). Recently, conformal prediction has also been extended to accommodate general statistical objects, such as graphs and functions that evolve over time, which can be very relevant in many biological problems \citep{matabuena2024conformal, lugosi2024uncertainty}.

The main contributions of our work are:


\begin{enumerate}

\item We propose and explore two conformal prediction algorithms for dynamical systems, specifically designed to optimize statistical efficiency when the measurement error of the biological system is homoscedastic. Specifically:
\begin{itemize}
    
\item The first algorithm focuses on achieving a specific quantile of calibration for each dimension of the dynamical system, making it more flexible when the homoscedasticity assumption is violated in any given coordinate.

\item The second algorithm is specially designed for large dynamical systems. The core idea is to perform a global standardization of the residuals, deriving a global quantile of calibration to return the final prediction regions.
\end{itemize}

\item We illustrate both algorithms using several case studies of increasing complexity and evaluate their performance in terms of statistical efficiency and computation time respect to traditional uncertainity quantification Bayesian methods.

\end{enumerate}


\section{Methodology}
\subsection*{Modeling framework and notation}
We consider dynamic models described by deterministic nonlinear ordinary differential equations (ODEs),
\begin{equation}\label{eq:model}
    \begin{aligned}
        \dot{x}(t) &= f(x,\theta,t ,x(t_0)) \\
        x(t_0) &= x_{0} \\
        y(t) &= g(x(t),\theta,t),
    \end{aligned}
\end{equation}

\noindent in which $x(t) \in \mathbb{R}^{n_x}$ is the vector of state variables at time $t$, $y(t) \in \mathbb{R}^{n_y}$ is the vector of observables at time $t$, and $\theta \in \mathbb{R}^{n_\theta}$ is the vector of unknown parameters. 

The vector field $f: \mathbb{R}^{n_x} \times \mathbb{R}^{n_\theta} \times \mathbb{R} \mapsto \mathbb{R}^{n_x}$ and the mappings $g: \mathbb{R}^{n_x} \times \mathbb{R}^{n_\theta} \times \mathbb{R} \mapsto \mathbb{R}^{n_y}$ and $x_0: \mathbb{R}^{n_\theta} \mapsto \mathbb{R}^{n_x}$ are possibly nonlinear.

\noindent The calibration of ODE models requires the estimation of parameter vector $\theta$ from theoretical observations of the output $y(t)$ at $n$ times, denoted as $t_1, t_2, \dots, t_{n}$. The total number of measurements is $n \times n_{y}$. In practice, we observe a perturbed multidimensional random variable \\
\\
$\tilde{y}= \begin{pmatrix}
\tilde{y}_{1,1} & \tilde{y}_{1,2} & \cdots & \tilde{y}_{1,n} \\
\tilde{y}_{2,1} & \tilde{y}_{2,2} & \cdots & \tilde{y}_{2,n} \\
\vdots & \vdots & \ddots & \vdots \\
\tilde{y}_{n_y,1} & \tilde{y}_{n_y,2} & \cdots & \tilde{y}_{n_y,n}
\end{pmatrix}$, and maybe we apply over $\tilde{y}$ a specific data transformation to convert the underlying model in homocedasticity in the transformated space. To be more precise, each random observation from the random vector $\tilde{y}$, as defined through the probabilistic model introduced below:

\begin{equation}\label{eq:noise}
\begin{split}
    h_{k}(\tilde{y}_{k,i},\lambda) = h_{k}(y_k(t_i),\lambda) + \epsilon_{k}(t_i)= h_{k}(g_{k}(x(t_i),\theta),\lambda) + \epsilon_{k,i}, \\  i = 1, \dots, n; k = 1, \dots, n_y, \lambda\in \mathbb{R}^{n_y},
\end{split}
\end{equation}

\noindent where, for each coordinate of dynamical system \( k = 1, \dots, n_y \), \(\epsilon_k \in \mathbb{R}^{n}\) denotes the measurement random noise of mean equal to zero, and \( h_k(\cdot, \lambda) \) is the specific transformation function that is an increasing real-valued function depending on a shape parameter \(\lambda\). Trivial examples of such functions include the logarithmic function, \(\log(\cdot)\), which is commonly used in the dynamical systems literature due to its equivalence with the log-normal probabilistic model \citep{Murphy2024}.
For identifiability purposes, model (\ref{eq:noise}) assumes that the variance of the random error \(\tilde{y}_{k,i}\) is solely a function of the regression function \( g_k(x(t_i), \theta)\). This assumption encompasses specific heteroscedastic cases, such as the log-normal model mentioned earlier, where the signal-noise ratio is homoscedastic in the transformed space but not in the original space.

\noindent   Model (\ref{eq:noise}) is a generalization of the Box-Cox transformation models, adapted for dynamical systems, and is defined in the regression literature by \cite{carroll1984power}. The authors refer to this semi-parametric transformation family as the transform-both-sides (TBS) model.

For simplicity, we assume that the measurement random noise is independent across the different temporal points $t_i$ $i=1,\dots, n$,  follows a normal distribution,  and is homoscedastic across dimensions, i.e., $\epsilon_{k,i} = \epsilon_{k}(t_i) \sim \mathcal{N}(0, \sigma_{k}^2),$ where $k = 1, \dots, n_y$. $\sigma_k(t_i)$ denotes the standard deviation for $k$ state of the dynamical systems in the temporal instant $t_i$. Here, we also assume that \( h_{k}(s,\lambda) = s \) for all \( s \in \mathbb{R} \), meaning that \( h_k \) is the identity function, or that we are  modeling the ODE systems directly from the original observations collected from the dynamical system. For simplicity, throughout this manuscript, we present all the modelling steps of the algorithms directly in terms of the regression function \( g(\cdot, \cdot) \) to aid the reader. The transformed version of the different algorithms introduced here only involves applying the specified data transformation to the original sample \(\overline{y}\) and then running the non-transformed algorithm in the transformed (or image) space.


\noindent The maximum likelihood estimate (MLE) of the vector of unknown parameters $\theta$, denoted as $\widehat{\theta}$, for any dataset $\mathcal{D}_n$ can be found by minimizing the log-likelihood function,

\begin{equation}\label{eq:llk}
\widehat{\theta}= \argmin_{\theta \in \mathbb{R}^{n_\theta}} \mathcal{L}(\theta;\mathcal{D}_n) = \frac{1}{2} \sum^{n_y}_{k=1} \sum^{n}_{i=1} \left[ \log(2\pi \sigma_{k}^2) + \left( \frac{\tilde{y}_{k,i} - g_{k}(x(t_i),\theta)}{\sigma_k} \right)^2 \right],
\end{equation}

\noindent which is the predominant optimization approach in the field of dynamical biological systems. Another popular approach, when no external information about the probabilistic mechanism of the random error \(\epsilon\) is available, is to use the minimization of the mean squared error as the optimization criterion to estimate the parameters of the regression function \(g\).

\subsection{Conformal prediction for dynamical systems}

Here, our objective is to present the development of new algorithms of conformal predictors for the class of dynamical systems stated above. The key idea is to consider the solution of the dynamical systems as the regression function \( g \) (for example, the conditional mean function) and the biological signals observed as corrupted by a measurement error \( \epsilon \). Then, using the residuals, it is possible to derive prediction regions using different conformal inference strategies. The underlying challenge in this type of regression is that the signal of the time series is observed in only a few \(n\) observations. Using full conformal methods, which require fitting hundreds or even thousands of models with \(n+1\) observations, is not advisable due to the prohibitive computational cost. This cost arises from the need to obtain model parameters through the mathematical optimization of large and high-dimensional systems of differential equations.

On the other hand, the split conformal methods (we fit each model component using random, disjoint splits) are not a good strategy from a statistical efficiency perspective when typically \( n < 20 \) in this type of biological problems. To mitigate the limitations of full and split conformal, we propose to use two new specific methods of conformal prediction for dynamical systems that increase statistical efficiency in different scenarios and are based on the application of jackknife techniques \citep{quenouille1956notes}.

Given a new random observation $Y_{n+1}= g(x(t_{n+1}),\theta)+\epsilon_{n+1}$, independent and identically distributed (i.i.d) with respect to \(\mathcal{D}_n\) (the original sample of $n$ observations collected from the dynamical system), and for a predefined confidence level \(\alpha \in [0,1]\), the goal of predictive inference is to provide a prediction region \(C^{\alpha}(\cdot) \subset \mathbb{R}^{n_y}\) such that \(\mathbb{P}(Y_{n+1} \in C^{\alpha}(t_{n+1})\mid T=t_{n+1}) = 1 - \alpha\), where the probability event is conditioned to the temporal instant $t_{n+1}$. We assume for practical purposes that such  prediction region exists and is unique.

Conformal prediction \citep{shafer2008tutorial,angelopoulos2023conformal}
 is a general uncertainty quantification framework that, independent of the underlying regression function \(g\) employed, provides non-asymptotic marginal (global) guarantees of the type \(\mathbb{P}(Y_{n+1} \in \widehat{C}^{\alpha}(t_{n+1})) \geq 1 - \alpha\). In this case, the probability \(\mathbb{P}\) is over the random sample \(\mathcal{D}_n \cup (t_{n+1},Y_{n+1})\). In the literature, there are three variants of conformal predictions related to the partitioning of the original random sample \(\mathcal{D}_n\) \citep{angelopoulos2023conformal,10.1214/20-AOS1965}: full, split, and jackknife conformal.

To determine whether a single prediction for the fixed pair \((t_{n+1}, Y_{n+1})\) is within the prediction region for any confidence level \(\alpha \in [0, 1]\), full conformal prediction requires using all \(n+1\) observations and estimating the regression model \(g\). For each fixed pair \((t_{n+1}, Y_{n+1})\), this involves optimizing the parameter vector \(\theta\) and evaluating whether the point lies inside or outside the prediction region by numerically approximating the predictions across a large-scale grid of observations, both in the temporal domain and in the states space. Such a procedure is computationally intensive, often requiring the estimation of hundreds of models, making it impractical for applications involving dynamical systems, especially in high-dimensional settings.

As a alternative, split conformal is valid for obtaining prediction regions for any new data point \((t_{n+1},Y_{n+1})\), and often divides the sample into $\lfloor n/2 \rfloor$ observations to estimate the function \(g\) and the remaining $n-\lfloor n/2 \rfloor$ to obtain the quantile for calibration to derive the final predictive regions.
Finally, the jackknife approach is an intermediate method for making predictions for any data point \((t_{n+1},Y_{n+1})\) and derive the prediction regions without sacrificing statistical efficiency. The jackknife approach requires the fitting of \(n\) predictive models, excluding in the \(i\)-th iteration the \(i^{\text{th}}\) observation. Here, in order to improve the robustness of the statistical properties of conformal jackknife, we derive two algorithms based on \cite{lei2018distribution} and the \texttt{Jackknife$+$}, a new jackknife conformal strategy proposed in \cite{10.1214/20-AOS1965}.

\subsection{Algorithms}
Algorithms \ref{alg:metd0} and \ref{alg:metd1} describe the core steps of our conformal UQ strategies for dynamical systems. In both algorithms, the first step involves excluding each $i$-th observation and fitting the regression functions to obtain the jackknife residuals. 

In the first algorithm, \texttt{CUQDyn1}, we apply a version of conformal \texttt{Jackknife$+$} in each coordinate of the dynamical system, introducing flexibility in the modeling in cases where uncertainty shape varies differently across each dimension. However, the theoretical convergence rates are often slower and require a larger number of observed data points in comparison with our second algorithm. 

To alleviate this issue and create a more efficient algorithm in some special homoscedastic cases, the second algorithm, \texttt{CUQDyn2}, uses the hypothesis that the model is homoscedastic along each coordinate. In its second step, we standardize the residuals using a prior estimation of the standard deviation. Then, we consider the global quantile of calibration. Finally, by re-scaling with the specific standard deviation, we obtain the final prediction interval.

\begin{algorithm}[ht!]
\caption{Conformal Naive UQ Algorithm for Dynamical Systems (\texttt{CUQDyn1})}
\begin{enumerate}
    \item For each $i=1, \ldots, n$, fit the regression model to the training data excluding the $i$th point to obtain $\widehat{g}_{-i}$. Compute the leave-one-out residual for each coordinate $k=1,\dots,n_{y}$ as $\epsilon_{i,k}=\left|y_{k}(t_i)-\widehat{g}_{-i,k}(t_{i})\right|$. Denote $\widehat{g}_{.,k}(t_{i})$ as the vector containing the estimates for state $k$ at time $t_i$ from all $\widehat{g}_{-i}$ fitted models, and denote $\epsilon_{.,k}$ as the vector of residuals for state $k$ across all time points $t_i$.
    
    \item For each coordinate $k=1,\dots,n_{y}$ and $i=1,\dots,n$, output the prediction interval:
    \begin{equation*}
        \left(q_\alpha\left(\widehat{g}_{.,k}(t_i)-\epsilon_{.,k}\right),\ q_{1-\alpha}\left(\widehat{g}_{.,k}(t_i)+\epsilon_{.,k}\right)\right),
    \end{equation*}
    where $\alpha$ is the predictive confidence level, and $q_\alpha$ and $q_{1-\alpha}$ denote the corresponding empirical quantiles for the levels $\alpha$ and $1-\alpha$, respectively.
\end{enumerate}
\label{alg:metd0}
\end{algorithm}

\begin{algorithm}[ht!]
    \caption{Conformal Global UQ Algorithm for Dynamical Systems (\texttt{CUQDyn2})}
\begin{enumerate}
    \item For each $i=1, \ldots, n$, fit the regression function $\widehat{g}_{-i}$ to the training data with the $i$th point removed, and compute the corresponding leave-one-out residual $\epsilon_{i,k}=\left|Y_{i}-\widehat{g}_{-i,k}(t_{i})\right|$, for $k=1,\dots,n_{y}$.
    
    \item For each coordinate $k=1,\dots,n_{y}$ and $i=1,\dots,n$, define the standardized variable $z_{i,k}=\frac{\epsilon_{i,k}}{\sigma_{k}}$, where $\sigma_{k}= \sqrt{\frac{1}{n}\sum_{j=1}^{n}\epsilon_{j,k}^{2}}$.
    
    \item Calculate the calibration quantile $q_{1-\alpha}$ for each coordinate $k=1,\dots,n_{y}$ using the sample $\{ z_{i,k} \}_{i=1}^{n}$.
    
    \item  Output the prediction interval for each coordinate $k$ as:
    \begin{equation*}
    \text{median}\left(\widehat{g}_{-1,k}(t_{i}), \dots, \widehat{g}_{-n,k}(t_{i})\right) \pm q_{1-\alpha}\sigma_{k}, \quad k=1,\dots,n_{y}.
    \end{equation*}
\end{enumerate}
\label{alg:metd1}
\end{algorithm}
Further implementation details (as a Matlab package) are given in the Supplementary Information.

\newpage

\subsection{Theory}

For this theorem, and all results that follow, all probabilities are stated with respect to the distribution of the training data points $\left(t_{1}, Y_{1}\right), \ldots,\left(t_{n}, Y_{n}\right)$ and the test data point $\left(t_{n+1}, Y_{n+1}\right)$ drawn i.i.d. from an arbitrary distribution $P$, and we assume implicitly that the regression method $g$ is invariant to the ordering of the data--invariant to permutations. We will treat the sample size $n \geq 2$ and the target coverage level $\alpha \in[0,1]$ as fixed throughout.

\textbf{Theorem:}    The conformal jacknife algorithms \texttt{CUQDyn1} and \texttt{CUQDyn2}  satisfies
$
\mathbb{P}(Y_{n+1} \in \widehat{C}^{\alpha}(t_{n+1}))
\} \geq 1-2 \alpha.
$

Consequence from \cite{10.1214/20-AOS1965}. The target in the inequality is $1-\alpha$ that is reached often except in some non-trivial mathematical examples.

\section{Results}
To assess the performance of the Uncertainty Quantification methods outlined in the previous section, we apply it to four case studies based on dynamic models of increasing complexity (Table 1 shows a summary): (i) a simple logistic growth model, (ii) a Lotka-Volterra classical model considering 2 species, (iii) the well-known $\alpha$-pinene kinetic model, widely used in parameter estimation studies, and (iv) the challenging NFKB signalling pathway. For each one of these models we formulated several parameter estimation problems generating synthetic data sets considering different scenarios for data cardinality and noise levels. All problems are fully observed except the NFKB case study. Due to space constraints, details about the models and the different estimation problems are given in the Supplementary Information.



Below, we present and discuss these case studies where we have considered synthetic datasets that have been generated considering a noise model described by Equation \eqref{eq:noise}. For simplicity, we assumed the errors to be normally distributed, centered around the noise-free data sample, and we have adopted a homoscedastic model, i.e. the variance remains constant across each dimension of the dataset. More specifically,
\begin{equation}\label{eq:nsnormal}
        \tilde{y}_{k,i} \sim \mathcal{N}(y_k(t_i), \sigma_{k}^2), \quad i = 1, \dots, n; \ k = 1, \dots, n_y,
\end{equation}
\noindent where, $\sigma_{k}=\epsilon\cdot\mu_k$, with $\mu_k=\sum_{i=1}^{n}y_k(t_i)/n$ capturing the mean value of state $k$, and $\epsilon$ representing the percentage of added noise. The ODE parameters $\theta$ are estimated by MLE equations for gaussian data as is specified in the Equation (\ref{eq:llk}).  Our results are compared with a Bayesian method (STAN), as described in the Supplementary Information.
\begin{table}
\caption{Summary of case study characteristics: number of unknown parameters ($n_\theta$), state variables ($n_x$), and measured observables ($n_y$)\label{tab1}}%
\begin{tabularx}{\columnwidth}{@{\extracolsep\fill}lXXX@{}}
\toprule
 & $n_\theta$  & $n_x$ & $n_y$\\
\midrule
Logistic    & 2 & 1 & 1  \\
Lotka-Volterra    & 4 & 2 & 2 \\
$\alpha$-Pinene    & 5 & 5  & 5  \\
NFKB  & 29 & 15 & 6 \\
\bottomrule
\end{tabularx}
\end{table}
\subsection{Case I: Logistic growth model}
To evaluate the performance of our methods on this case study, we considered various scenarios with different noise levels ($0\%$, $1\%$, $5\%$ and $10\%$) and dataset sizes (10, 20, 50 and 100 data points). For each combination of noise level and dataset size, we generated 50 different synthetic datasets, totaling 800 unique datasets. By generating multiple datasets for each scenario, we were able to obtain a robust estimate of the methods' behavior and assess their consistency across different realizations of the data.

The comparative analysis of the logistic growth model, as shown in Figure \ref{fig:log_ex}, highlights the robustness of the proposed methods \texttt{CUQDyn1} and \texttt{CUQDyn2} compared to conventional methodologies such as the Bayesian approach implemented with STAN. For a 10-point synthetic dataset with a 10 percent noise level, the predictive regions obtained by both conformal methods showed good coverage without requiring prior calibration of the models, unlike the Bayesian approach. Moreover, both \texttt{CUQDyn1} and \texttt{CUQDyn2} yield predictive regions comparable to those generated by the jackknife+ method; however, in this particular case, the \texttt{CUQDyn1} method shows superior performance.

In terms of computational efficiency, the conformal methods proved to be marginally faster than STAN, even for a problem of this small size, with differences on the order of a few seconds. This makes them more suitable for real-time applications.

\begin{figure}[H]
    \centering
    \scalebox{0.6}{
    \includegraphics{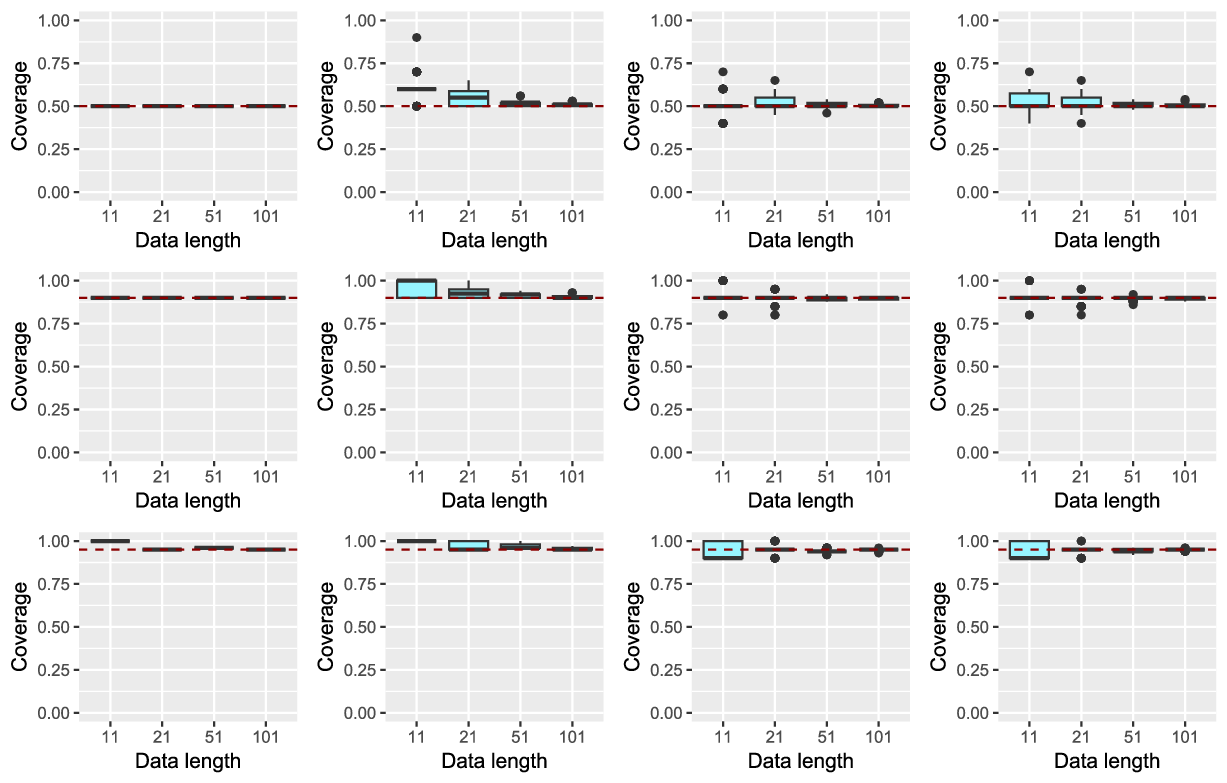}}
 \caption{
Boxplot of marginal coverage $\mathbb{P}(Y_{n+1} \in \widehat{C}^{\alpha}(X_{n+1}))$ for different sample sizes and $\alpha = 0.05$, $0.1$, and $0.5$ of our first algorithm, \texttt{CUQDyn1}, is presented for different noise levels ($0\%$, $1\%$, $5\%$, and $10\%$) across different columnss. The results remain very stable across all examined cases for larger sample sizes of 100 temporal points.}
    \label{fig:log-boxp}
\end{figure}

To examine the marginal coverage $\mathbb{P}(Y_{n+1} \in \widehat{C}^{\alpha}(X_{n+1}))$ for $\alpha=0.05, 0.1, 0.5$ of our first algorithm \texttt{CUQDyn1}, see Figure \ref{fig:log-boxp}
 for different signal noises and sample sizes. The figure indicates the good empirical performance of our algorithm, achieving the desired nominal level in expectation as is expected for the non-assympotic guarantees of conformal prediction methods.

\begin{figure}[H]
    \centering
    \scalebox{0.8}{
    \includegraphics{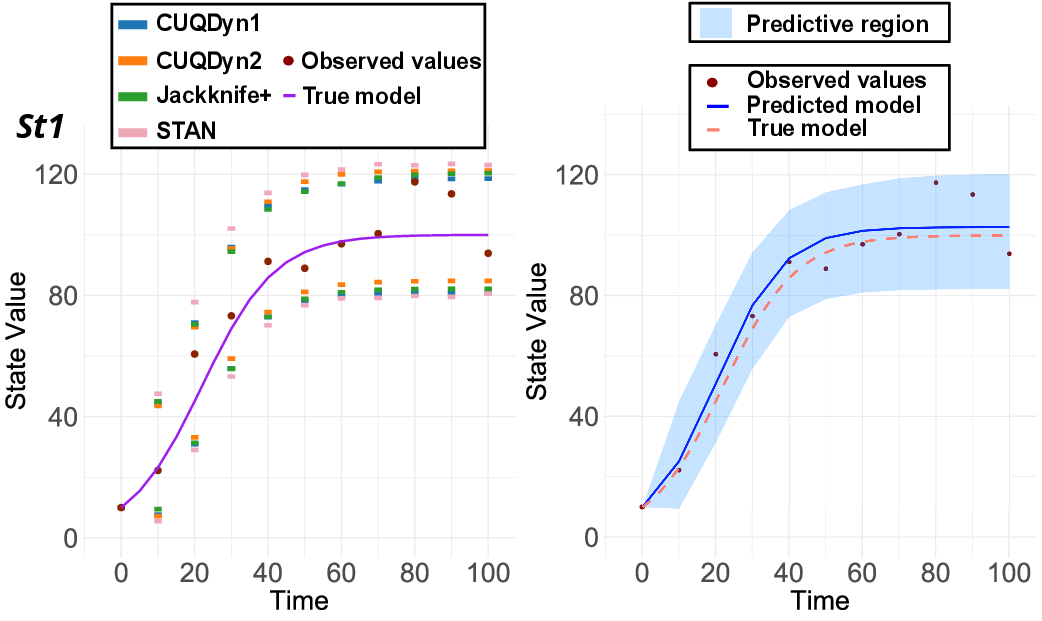}}
 \caption{Comparative analysis of the Logistic model predictive regions. This figure presents the $95\%$ predictive regions obtained from a 10-point dataset subjected to $10\%$ noise. The left subplot showcases results using four different methodologies: our two proposed methods (\texttt{CUQDyn1} and \texttt{CUQDyn2}), the original jackknife+ method and a Bayesian approach implemented with STAN. The right subplot shows the predictive region and the predicted model for the \texttt{CUQDyn1} algorithm applied to the same dataset. Numerical results related to this example are available in the Supplementary Information.}
    \label{fig:log_ex}
\end{figure}

\subsection{Case II: Lotka-Volterra model}
For this case study we generated datasets with the same noise levels ($0\%$, $1\%$, $5\%$ and $10\%$) as in the previous example and three different sizes (30, 60 and 120 points). Additionally, for each combination of noise level and dataset size, we generated 50 different synthetic datasets, resulting in a total of 600 unique datasets. 

Figure \ref{fig:lv_ex}
 shows the results in a 30-point Lotka-Volterra
dataset, indicating that the predictive regions generated by
the conformal methods and STAN are similar in terms of
coverage. However, as in the previous case, \texttt{CUQDyn1} and \texttt{CUQDyn2}
offer the advantage of not requiring extensive hyperparameter
tuning, while also being more computationally efficient. In this
particular example, while the bayesian method obtains results
within a timeframe on the order of minutes, both conformal
methods achieve this in a significantly shorter span, on the order of seconds.
\begin{figure}[ht!]
    \centering
    \scalebox{0.8}{
    \includegraphics{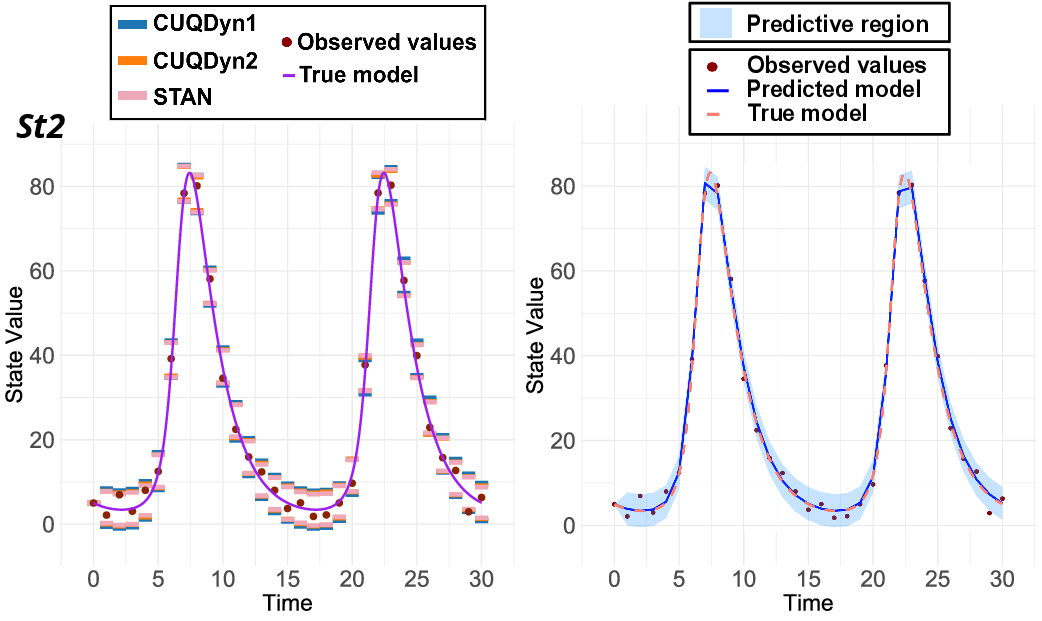}}
    \caption{Comparative analysis of the Lotka-Volterra model predictive regions for the second state. This figure presents the $95\%$ predictive regions obtained from a 30-point dataset subjected to $10\%$ noise. The left subplot showcase results using three different methodologies: our two proposed methods (\texttt{CUQDyn1} and \texttt{CUQDyn2}) and a Bayesian approach implemented with STAN. The right subplot shows the predictive region and the predicted model for the \texttt{CUQDyn2} algorithm applied to the same dataset. Numerical results related to this example are available in the Supplementary Information.
    \\}
    \label{fig:lv_ex}
\end{figure}
\subsection{Case III: Isomerization of $\alpha$-Pinene}

The dataset generation procedure for this case study mirrored that used for the Logistic model, employing the same noise levels and dataset sizes. Although we generated synthetic datasets to assess the method's behavior, we illustrated this behavior with a real dataset from \cite{box1973some}.

Figure \ref{fig:alpha}
 shows the resulting regions of the isomerization of $\alpha$-Pinene by applying the different algorithms to the 9-point real dataset. The results are once again consistent between both conformal algorithms and closely align with the regions obtained using STAN. In terms of computational cost, the conformal algorithms are notably more efficient, requiring less than a minute to compute the regions, whereas the Bayesian approach takes several minutes.

\begin{figure}[H]
    \centering
    \scalebox{0.8}{
    \includegraphics{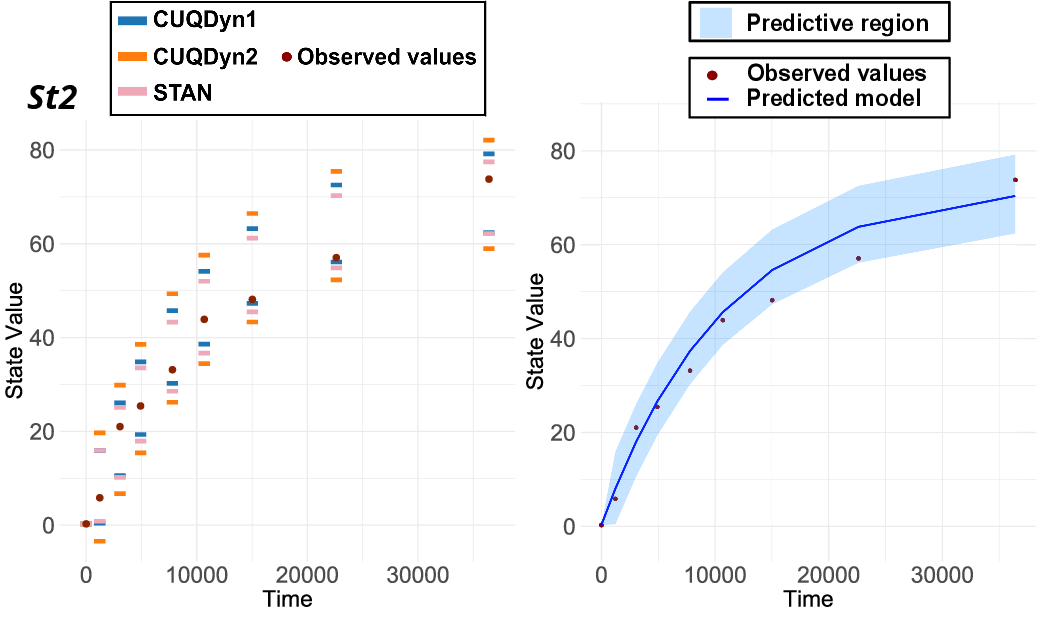}}

    \caption{Comparative analysis of the $\alpha$-pinene isomerization model predictive regions for the second state. This figure presents the $95\%$ predictive regions obtained from a 9-point real dataset. The left subplot showcases results using three different methodologies: our two proposed methods (\texttt{CUQDyn1} and \texttt{CUQDyn2}) and a Bayesian approach implemented with STAN. The right subplot shows the predictive region and the predicted model for the \texttt{CUQDyn1} algorithm applied to the same dataset. Numerical results related to this example are available in the Supplementary Information.}
    \label{fig:alpha}
\end{figure}

\subsection{Case IV: NFKB signaling pathway}

The dataset generation process was similar to the one used in the previous case studies. The results of applying different algorithms to a 13-point synthetic dataset are illustrated in the Supplementary Information. Both conformal inference methods yielded accurate and consistent results in short computation times. In contrast, STAN failed to converge, even after very long computations. This non-convergence is likely due to partial lack of identifiability.

\section{Discussion}

In this study, we present two algorithms, \texttt{CUQDyn1} and \texttt{CUQDyn2}, using conformal methods to perform uncertainty quantification. These methods allow the computation of prediction regions for nonlinear dynamic models of biological systems. We assume that the signal-to-noise ratio is homoscedastic in the measurements or any transformation of the original data. We successfully compared the performance of these new methods with Bayesian approaches using a set of problems of increasing complexity. The main conclusions from the numerical results are summarized below.

Our algorithms were significantly faster than the Bayesian method (STAN) for the case studies examined. Our methods, which do not require tuning of hyperparameters, performed well in agreement with the Bayesian approach for smaller case studies and larger datasets (more than 50 temporal points). However, for high-dimensional biological systems, as illustrated with the NFKB case study, our conformal methods exhibited better accuracy, while STAN encountered convergence issues.

Furthermore, our methods, although not based on specified regression models, achieved good marginal coverage due to their non-asymptotic properties. In contrast, Bayesian methods, exemplified by STAN, showed a more critical impact of poor calibration on marginal coverage, especially for small sample sizes, due to the lack of non-asymptotic properties.

Our study also revealed that obtaining good coverage properties with a Bayesian method requires careful tuning of the prior, which can be challenging even for well-known small problems and may be very difficult for new, larger problems arising in real applications. We encountered convergence issues with the MCMC strategy in STAN, likely due to the multimodal nature of posterior distributions and identifiability issues, consistent with previous reports \citep{raue2013joining}.

A primary limitation of our new methods is that the prediction regions might occasionally take negative values when observed states are very close to zero, which may be unrealistic from a mechanistic perspective. This issue is observed in both the STAN Bayesian implementation and the conformal methods presented here. One potential cause is the assumption of homoscedasticity (i.e., consistent signal-to-noise ratio across the entire domain). Moreover, the underlying model may not be correctly specified across all parts of the domain.

A straightforward solution to this issue is to apply a prior data transformation, such as the log-normal transformation, which allows for modeling heteroscedastic scenarios, or a more general family of possible transformations introduced in the model \eqref{eq:noise}. To enhance the usability of our proposed methods within the scientific community, we have made the code and data from this study available in a public repository. Moving forward, we plan to offer updates, including optimized code for high-performance computers, novel validations, and automatic transformation approaches for various error structures in the modeling.

Overall, our study presents a new framework for uncertainty quantification in dynamical models using conformal prediction methods. This framework provides an alternative to classical Bayesian methods. Notably, our new methods are computationally scalable, which is crucial for large biological models. From a mathematical statistics perspective, they offer non-asymptotic guarantees and avoid the technical difficulties of calibrating prior functions necessary in Bayesian statistics. For future work, we suggest exploring conformal quantile algorithms for massive dynamical biological systems \citep{romano2019conformalized}. These algorithms typically provide better conditional coverage than other conformal algorithms \citep{sesia2020comparison} and do not require the assumption of symmetric random errors. However, applying quantile conformal algorithms in practice may require collecting more temporal observations of dynamical systems, which might not always be feasible in real-world scenarios.

\section*{Acknowledgments}
JRB acknowledges support from grant PID2020-117271RB-C22 (BIODYNAMICS) funded by MCIN/AEI/10.13039/501100011033, 
from grant PID2023-146275NB-C22 (DYNAMO-bio) funded by MICIU/AEI/ 10.13039/501100011033 and ERDF/EU, and from grant CSIC PIE 202470E108 (LARGO). The authors also wish to thank Javier Enrique Aguilar Romero for his assistance with the use of Stan. The funders had no role in the study design, data collection and analysis, decision to publish, or preparation of the manuscript




\newpage

\begin{appendix}
\renewcommand{\thesection}{\Alph{section}}
\section*{APPENDIX - Supplementary material}

\section{Matlab implementation of the algorithms}

We implemented our \texttt{CUQDyn1} and \texttt{CUQDyn2} algorithms in Matlab. Parameter estimations were formulated as the minimization of a least squares cost function subject the dynamics (described by the model ODEs) and parameter bounds. These problems are non-convex and were solved using a global hybrid method, enhanced scatter search (\texttt{eSS}) due to its good performance an robustness \citep{villaverde2018benchmarking}. \texttt{eSS} is available in Matlab as part of the \texttt{MEIGO} optimization toolbox \citep{egea2014meigo}. Our code also has dependencies with the the Optimization Toolbox and the Parallel Computing Toolbox.  

The software for the methodology and the reproduction of the results is available at \href{https://zenodo.org/doi/10.5281/zenodo.13644869}{https://zenodo.org/doi/10.5281/zenodo.13644869}. All computations were carried out on a PC DELL Precision 7920 workstation with dual Intel Xeon Silver 4210R processors.

\section{Comparison with a Bayesian method}

Bayesian methods are a classical approach for performing automated uncertainty quantifications by estimating the posterior distribution \(P(\theta\mid\mathcal{D}_n)\), where \(\theta\) represents the parameter of interest and \(\mathcal{D}_n = \{X_i\}_{i=1}^n\) denotes the observed data. The key components in Bayesian analysis are the prior distribution \(P(\theta)\), which encapsulates our initial beliefs about \(\theta\), and the likelihood function \(P(\mathcal{D}_n \mid \theta)\), which represents the probability of observing the data \(\mathcal{D}_n\) given the parameter \(\theta\).

In many practical scenarios, computing the posterior distribution analytically is challenging. Markov Chain provide Monte Carlo (MCMC) methods provides a general and powerful techniques used to estimate the posterior distribution by generating samples from it. Notable MCMC algorithms include Metropolis-Hastings and Gibbs sampling.

Nowadays, there are general software tools available for implementing Bayesian inference and MCMC methods. One such tool is STAN. To use STAN, one writes a model in the STAN modeling language, which involves defining the data, parameters, and model (i.e., prior and likelihood). STAN can be seamlessly integrated with \texttt{R} through the \texttt{rstan} package \cite{guo2020package}, allowing users to perform Bayesian analyses within the R environment. The \texttt{rstan} package provides functions to compile STAN models, fit them to data, and extract samples for posterior analysis. Our implementations of the different case studies are also available in the Zenodo link above.

\section{Case studies}




\subsection{Case I: Logistic growth model}

As our initial case study we considered the well-known logistic model \citep{Tsoularis2002}, governed by a single differential equation with two unknown parameters. This model is frequently used in population growth and epidemic spread modeling.

\begin{equation}\label{eq:log_eqs}
        \dot{x} = rx\left(1-\frac{x}{K}\right).
\end{equation}

Here, $r$ represents the growth rate, and $K$ denotes the carrying capacity. The initial condition considered in the generation of the datasets was $x(0)=10$. Additionally, the values of the parameters used were $r=0.1$ and $K=100$. The initial condition is assumed to be known across all case studies considered. Since this logistic model has an analytical solution, it facilitates the comparison of our methods' performance with other established conformal methods for algebraic models, such as the jackknife+ \citep{10.1214/20-AOS1965}.

To evaluate the performance of our methods on this case study, we considered various scenarios with different noise levels ($0\%$, $1\%$, $5\%$ and $10\%$) and dataset sizes (10, 20, 50 and 100 data points). For each combination of noise level and dataset size, we generated 50 different synthetic datasets, totaling 800 unique datasets. By generating multiple datasets for each scenario, we were able to obtain a robust estimate of the methods' behavior and assess their consistency across different realizations of the data.

The comparative analysis of the logistic growth model, as shown in Figure \ref{fig:log_ex}, highlights the robustness of the proposed methods \texttt{CUQDyn1} and \texttt{CUQDyn2} compared to conventional methodologies such as the Bayesian approach implemented with STAN. For a 10-point synthetic dataset with a 10 percent noise level, the predictive regions obtained by both conformal methods showed good coverage without requiring prior calibration of the models, unlike the Bayesian approach. Moreover, both \texttt{CUQDyn1} and \texttt{CUQDyn2} yield predictive regions comparable to those generated by the jackknife+ method; however, in this particular case, the \texttt{CUQDyn1} method shows superior performance.

In terms of computational efficiency, the conformal methods proved to be marginally faster than STAN, even for a problem of this small size, with differences on the order of a few seconds. This makes them more suitable for real-time applications.

To examine the marginal coverage $\mathbb{P}(Y_{n+1} \in \widehat{C}^{\alpha}(X_{n+1}))$ for $\alpha=0.05, 0.1, 0.5$ of our first algorithm \texttt{CUQDyn1}, see Figure \ref{fig:log-boxp}
 for different signal noises and sample sizes. The figure indicates the good empirical performance of our algorithm, achieving the desired nominal level in expectation.

\begin{figure}
    \centering
    \scalebox{0.7}{
    \includegraphics{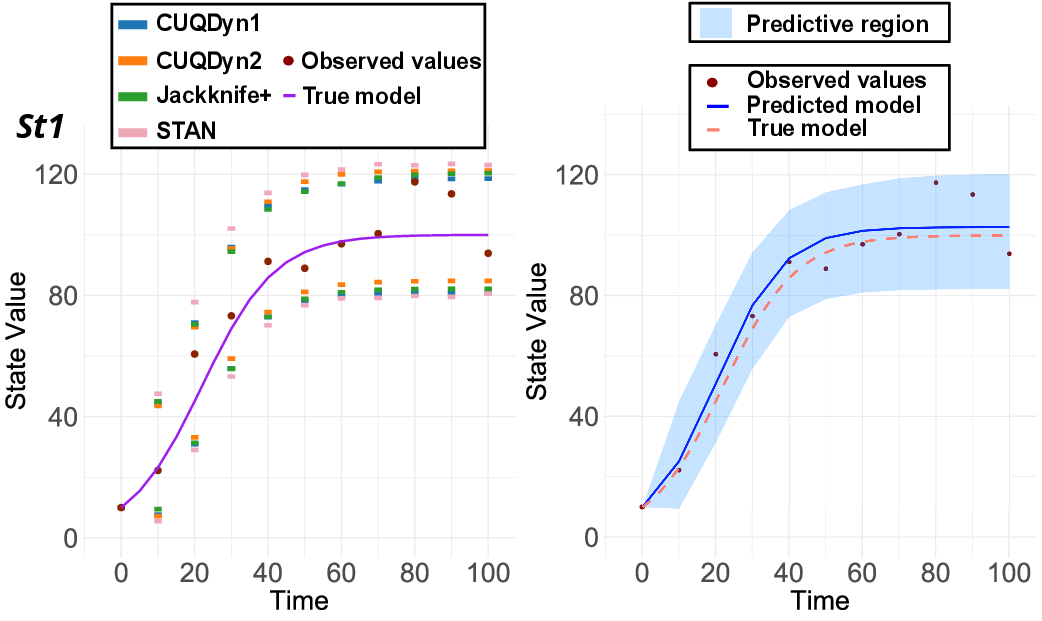}}
 \caption{Comparative analysis of the Logistic model predictive regions. This figure presents the $95\%$ predictive regions obtained from a 10-point dataset subjected to $10\%$ noise. The left subplot showcases results using four different methodologies: our two proposed methods (\texttt{CUQDyn1} and \texttt{CUQDyn2}), the original jackknife+ method and a Bayesian approach implemented with STAN. The right subplot shows the predictive region and the predicted model for the \texttt{CUQDyn1} algorithm applied to the same dataset.}
    \label{fig:log_ex}
\end{figure}

\begin{sidewaystable*}[!htbp]
\caption{Numerical results corresponding to the comparative analysis of the Logistic model predictive regions. This table presents the $95\%$ lower and upper predictive bounds (LPB and UPB, respectively) obtained for each time point ($t$) in a 10-point dataset subjected to $10\%$ noise. The observed data ($y$) and the true state value ($x_{nom}$) are also shown. The results are reported for four methodologies: the proposed \texttt{CUQDyn1} and \texttt{CUQDyn2} methods, the original jackknife+ approach, and a Bayesian method implemented with STAN.\label{tab:log-comp}}
\setlength{\tabcolsep}{1.9pt}
\begin{tabular*}{\textwidth}{@{\extracolsep{\fill}}lccccccccccc}
\toprule
& & & \multicolumn{2}{c}{CUQDyn1} & \multicolumn{2}{c}{CUQDyn2} & \multicolumn{2}{c}{Jackknife+} & \multicolumn{2}{c}{STAN}\\
\cline{4-5}\cline{6-7}\cline{8-9}\cline{10-11}
$t$ & $y$ & $x_{nom}$ & LPB & UPB & LPB & UPB & LPB & UPB & LPB & UPB\\
\midrule
0 & 1.000e+01 & 1.000e+01 & 1.000e+01 & 1.000e+01 & 1.000e+01 & 1.000e+01 & 1.000e+01 & 1.000e+01 & 1.000e+01 & 1.000e+01 & \\
10 & 2.226e+01 & 2.320e+01 & 7.698e+00 & 4.440e+01 & 7.213e+00 & 4.357e+01 & 9.493e+00 & 4.507e+01 & 5.545e+00 & 4.756e+01 & \\
20 & 6.066e+01 & 4.509e+01 & 3.092e+01 & 7.100e+01 & 3.318e+01 & 6.954e+01 & 3.140e+01 & 7.043e+01 & 2.904e+01 & 7.781e+01 & \\
30 & 7.327e+01 & 6.906e+01 & 5.593e+01 & 9.594e+01 & 5.916e+01 & 9.552e+01 & 5.578e+01 & 9.442e+01 & 5.324e+01 & 1.021e+02 & \\
40 & 9.123e+01 & 8.585e+01 & 7.284e+01 & 1.095e+02 & 7.453e+01 & 1.109e+02 & 7.296e+01 & 1.083e+02 & 7.015e+01 & 1.138e+02 & \\
50 & 8.895e+01 & 9.428e+01 & 7.816e+01 & 1.149e+02 & 8.116e+01 & 1.175e+02 & 7.888e+01 & 1.143e+02 & 7.681e+01 & 1.198e+02 & \\
60 & 9.703e+01 & 9.782e+01 & 7.997e+01 & 1.167e+02 & 8.357e+01 & 1.199e+02 & 8.105e+01 & 1.169e+02 & 7.905e+01 & 1.215e+02 & \\
70 & 1.004e+02 & 9.919e+01 & 8.055e+01 & 1.177e+02 & 8.441e+01 & 1.208e+02 & 8.181e+01 & 1.189e+02 & 7.917e+01 & 1.232e+02 & \\
80 & 1.174e+02 & 9.970e+01 & 8.073e+01 & 1.183e+02 & 8.468e+01 & 1.210e+02 & 8.207e+01 & 1.198e+02 & 7.982e+01 & 1.229e+02 & \\
90 & 1.135e+02 & 9.989e+01 & 8.079e+01 & 1.185e+02 & 8.477e+01 & 1.211e+02 & 8.215e+01 & 1.201e+02 & 7.953e+01 & 1.234e+02 & \\
100 & 9.390e+01 & 9.996e+01 & 8.081e+01 & 1.185e+02 & 8.480e+01 & 1.212e+02 & 8.218e+01 & 1.203e+02 & 8.060e+01 & 1.230e+02 & \\
\bottomrule
\end{tabular*}
\end{sidewaystable*}

\subsection{Case II: Lotka-Volterra model}

As a second case study, we considered a two species Lotka-Volterra model \citep{wangersky1978lotka}, often referred to as the predator-prey model. This model provides a fundamental framework for studying the dynamics between two interacting species. In its simplest form, it describes the interactions between a predator species and a prey species through a set of coupled differential equations with four unknown parameters:

\begin{equation}\label{eq:lv_eqs}
    \begin{aligned}
        \dot{x}_1 &= x_1(\alpha-\beta x_2), \\
        \dot{x}_2 &= -x_2(\gamma-\delta x_1).
    \end{aligned}
\end{equation}

Here, $x_1$ and $x_2$ represent the populations of the prey and predator, respectively. The parameters $\alpha$, $\beta$, $\gamma$ and $\delta$ are positive constants representing the interactions between the two species. Specifically, this parameters dictate the growth rates and interaction strengths, capturing the essence of biological interactions such as predation and competition. The initial conditions considered in the generation of the datasets were $x(0)=(10,5)$. Additionally, the values of the parameters used were $\alpha=\gamma=0.5$ and $\beta=\delta=0.02$.

For this case study we generated datasets with the same noise levels ($0\%$, $1\%$, $5\%$ and $10\%$) as in the previous example and three different sizes (30, 60 and 120 points). Additionally, for each combination of noise level and dataset size, we generated 50 different synthetic datasets, resulting in a total of 600 unique datasets. 

Figure \ref{fig:lv_ex}
 shows the results in a 30-point Lotka-Volterra dataset, indicating that the predictive regions generated by the conformal methods and STAN are similar in terms of coverage. However, as in the previous case, \texttt{CUQDyn1} and \texttt{CUQDyn2} offer the advantage of not requiring extensive hyperparameter tuning, while also being more computationally efficient. In this particular example, while the bayesian method obtains results within a timeframe on the order of minutes, both conformal methods achieve this in a significantly shorter span, on the order of seconds.

\begin{figure}
    \centering
    \scalebox{0.7}{
    \includegraphics{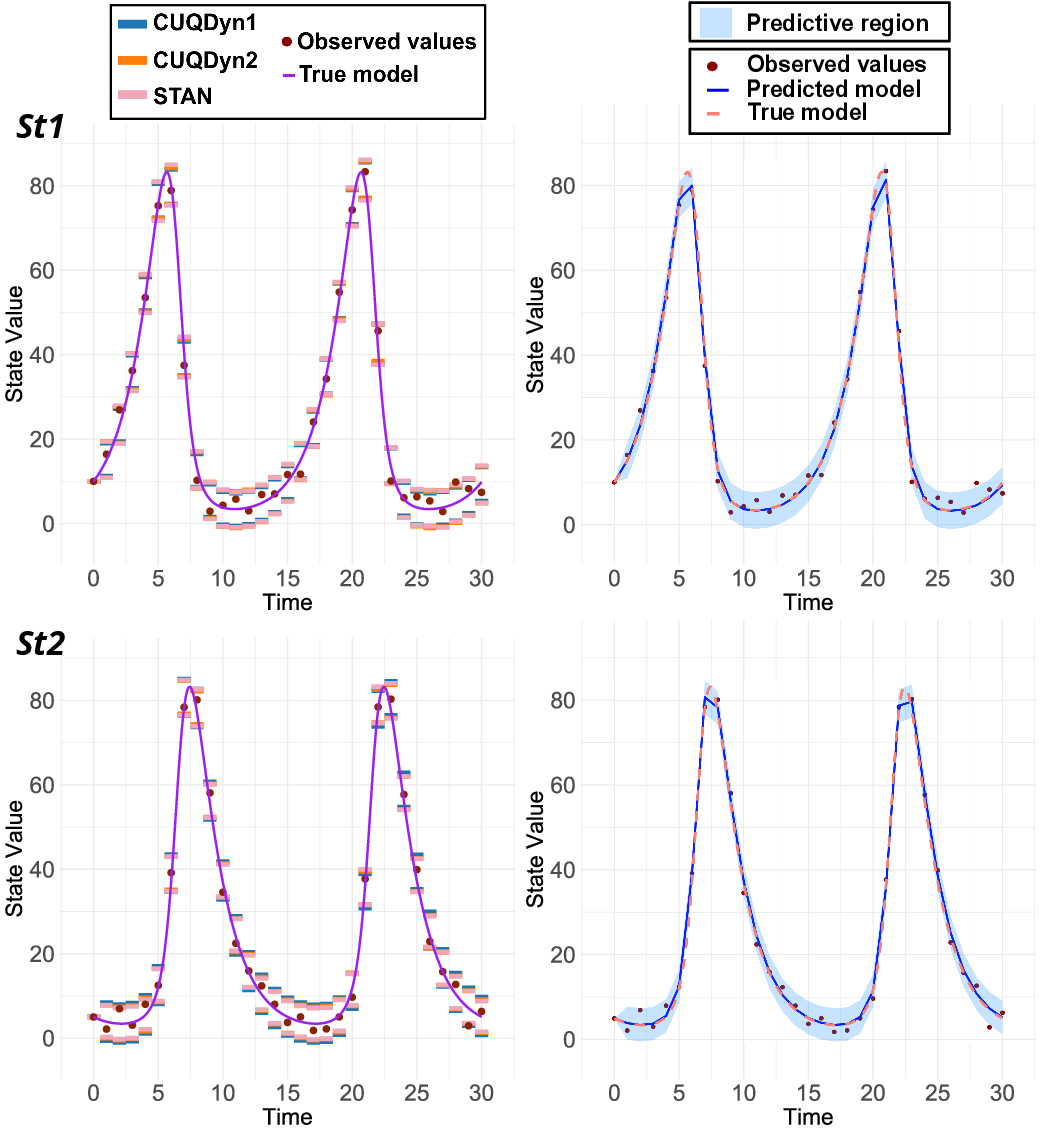}}
    \caption{Comparative analysis of the Lotka-Volterra model predictive regions. This figure presents the $95\%$ predictive regions obtained from a 30-point dataset subjected to $10\%$ noise. The two left subplots showcase results using three different methodologies: our two proposed methods (\texttt{CUQDyn1} and \texttt{CUQDyn2}) and a Bayesian approach implemented with STAN. The two right subplots show the predictive region and the predicted model for the \texttt{CUQDyn2} algorithm applied to the same dataset.}
    \label{fig:lv_ex}
\end{figure}

\begin{table*}[!htbp]
\caption{Numerical results corresponding to the comparative analysis of the Lotka-Volterra model predictive regions for the first state. This table presents the $95\%$ lower and upper predictive bounds (LPB and UPB, respectively) obtained for each time point ($t$) in a 30-point dataset subjected to $10\%$ noise. The observed data ($y$) and the true state value ($x_{nom}$) are also shown. The results are reported for three methodologies: the proposed \texttt{CUQDyn1} and \texttt{CUQDyn2} methods and a Bayesian method implemented with STAN.\label{tab:lv-comp-1}}
\setlength{\tabcolsep}{1.9pt}
\begin{tabular*}{\textwidth}{@{\extracolsep{\fill}}lccccccccc}
\toprule
& & & \multicolumn{2}{c}{CUQDyn1} & \multicolumn{2}{c}{CUQDyn2} & \multicolumn{2}{c}{STAN}\\
\cline{4-5}\cline{6-7}\cline{8-9}
$t$ & $y$ & $x_{nom}$ & LPB & UPB & LPB & UPB & LPB & UPB \\
\midrule
0 & 1.000e+01 & 1.000e+01 & 1.000e+01 & 1.000e+01 & 1.000e+01 & 1.000e+01 & 1.000e+01 & 1.000e+01 & \\
1 & 1.627e+01 & 1.510e+01 & 9.974e+00 & 2.025e+01 & 9.535e+00 & 2.068e+01 & 1.022e+01 & 1.984e+01 & \\
2 & 2.597e+01 & 2.317e+01 & 1.804e+01 & 2.831e+01 & 1.760e+01 & 2.874e+01 & 1.832e+01 & 2.832e+01 & \\
3 & 3.373e+01 & 3.562e+01 & 3.046e+01 & 4.073e+01 & 3.000e+01 & 4.114e+01 & 3.069e+01 & 4.046e+01 & \\
4 & 5.374e+01 & 5.379e+01 & 4.854e+01 & 5.882e+01 & 4.807e+01 & 5.921e+01 & 4.837e+01 & 5.925e+01 & \\
5 & 7.607e+01 & 7.551e+01 & 7.009e+01 & 8.036e+01 & 6.961e+01 & 8.075e+01 & 6.965e+01 & 8.039e+01 & \\
6 & 7.493e+01 & 8.002e+01 & 7.344e+01 & 8.429e+01 & 7.304e+01 & 8.418e+01 & 7.317e+01 & 8.386e+01 & \\
7 & 4.005e+01 & 4.053e+01 & 3.381e+01 & 4.477e+01 & 3.347e+01 & 4.461e+01 & 3.364e+01 & 4.447e+01 & \\
8 & 1.435e+01 & 1.316e+01 & 7.461e+00 & 1.773e+01 & 7.070e+00 & 1.821e+01 & 7.415e+00 & 1.762e+01 & \\
9 & 6.658e+00 & 5.733e+00 & 2.295e-01 & 1.050e+01 & -1.836e-01 & 1.096e+01 & 6.027e-01 & 1.044e+01 & \\
10 & 1.580e+00 & 3.786e+00 & -1.595e+00 & 8.677e+00 & -2.018e+00 & 9.122e+00 & -1.310e+00 & 8.689e+00 & \\
11 & 6.848e+00 & 3.424e+00 & -1.915e+00 & 8.357e+00 & -2.340e+00 & 8.800e+00 & -1.494e+00 & 8.226e+00 & \\
12 & 6.422e+00 & 3.821e+00 & -1.528e+00 & 8.744e+00 & -1.953e+00 & 9.187e+00 & -1.257e+00 & 8.470e+00 & \\
13 & 4.250e+00 & 4.878e+00 & -5.077e-01 & 9.764e+00 & -9.314e-01 & 1.021e+01 & -1.349e-01 & 9.607e+00 & \\
14 & 4.422e+00 & 6.777e+00 & 1.309e+00 & 1.158e+01 & 8.889e-01 & 1.203e+01 & 1.878e+00 & 1.148e+01 & \\
15 & 8.831e+00 & 9.921e+00 & 4.314e+00 & 1.459e+01 & 3.899e+00 & 1.504e+01 & 4.873e+00 & 1.474e+01 & \\
16 & 1.784e+01 & 1.498e+01 & 9.152e+00 & 1.942e+01 & 8.744e+00 & 1.988e+01 & 9.879e+00 & 1.965e+01 & \\
17 & 2.148e+01 & 2.297e+01 & 1.681e+01 & 2.708e+01 & 1.641e+01 & 2.755e+01 & 1.756e+01 & 2.741e+01 & \\
18 & 3.687e+01 & 3.532e+01 & 2.866e+01 & 3.893e+01 & 2.828e+01 & 3.942e+01 & 2.959e+01 & 3.937e+01 & \\
19 & 5.202e+01 & 5.337e+01 & 4.619e+01 & 5.667e+01 & 4.582e+01 & 5.696e+01 & 4.771e+01 & 5.748e+01 & \\
20 & 7.378e+01 & 7.512e+01 & 6.812e+01 & 7.858e+01 & 6.774e+01 & 7.888e+01 & 6.851e+01 & 7.887e+01 & \\
21 & 8.134e+01 & 8.039e+01 & 7.673e+01 & 8.701e+01 & 7.624e+01 & 8.738e+01 & 7.399e+01 & 8.524e+01 & \\
22 & 4.315e+01 & 4.139e+01 & 3.909e+01 & 4.999e+01 & 3.918e+01 & 5.032e+01 & 3.616e+01 & 4.739e+01 & \\
23 & 1.409e+01 & 1.343e+01 & 8.884e+00 & 1.916e+01 & 8.355e+00 & 1.950e+01 & 8.841e+00 & 1.841e+01 & \\
24 & 1.085e+01 & 5.804e+00 & 5.953e-01 & 1.087e+01 & 1.348e-01 & 1.128e+01 & 1.022e+00 & 1.065e+01 & \\
25 & 4.580e+00 & 3.804e+00 & -1.537e+00 & 8.734e+00 & -1.974e+00 & 9.166e+00 & -1.097e+00 & 8.501e+00 & \\
26 & -2.494e+00 & 3.423e+00 & -1.973e+00 & 8.298e+00 & -2.405e+00 & 8.736e+00 & -1.860e+00 & 8.504e+00 & \\
27 & 1.040e+00 & 3.807e+00 & -1.669e+00 & 8.602e+00 & -2.098e+00 & 9.043e+00 & -1.317e+00 & 8.542e+00 & \\
28 & 1.479e+00 & 4.849e+00 & -7.498e-01 & 9.522e+00 & -1.177e+00 & 9.964e+00 & -2.282e-01 & 9.313e+00 & \\
29 & 1.030e+01 & 6.729e+00 & 9.260e-01 & 1.120e+01 & 5.132e-01 & 1.165e+01 & 1.641e+00 & 1.140e+01 & \\
30 & 9.117e+00 & 9.842e+00 & 3.710e+00 & 1.398e+01 & 3.310e+00 & 1.445e+01 & 4.792e+00 & 1.448e+01 & \\
\bottomrule
\end{tabular*}
\end{table*}

\begin{table*}[!htbp]
\caption{Numerical results corresponding to the comparative analysis of the Lotka-Volterra model predictive regions for the second state. This table presents the $95\%$ lower and upper predictive bounds (LPB and UPB, respectively) obtained for each time point ($t$) in a 30-point dataset subjected to $10\%$ noise. The observed data ($y$) and the true state value ($x_{nom}$) are also shown. The results are reported for three methodologies: the proposed \texttt{CUQDyn1} and \texttt{CUQDyn2} methods and a Bayesian method implemented with STAN.\label{tab:lv-comp-2}}
\setlength{\tabcolsep}{1.9pt}
\begin{tabular*}{\textwidth}{@{\extracolsep{\fill}}lccccccccc}
\toprule
& & & \multicolumn{2}{c}{CUQDyn1} & \multicolumn{2}{c}{CUQDyn2} & \multicolumn{2}{c}{STAN}\\
\cline{4-5}\cline{6-7}\cline{8-9}
$t$ & $y$ & $x_{nom}$ & LPB & UPB & LPB & UPB & LPB & UPB \\
\midrule
0 & 5.000e+00 & 5.000e+00 & 5.000e+00 & 5.000e+00 & 5.000e+00 & 5.000e+00 & 5.000e+00 & 5.000e+00 & \\
1 & -3.196e-02 & 3.883e+00 & -1.135e+00 & 8.952e+00 & -2.394e+00 & 1.019e+01 & -1.270e+00 & 8.999e+00 & \\
2 & 5.589e+00 & 3.432e+00 & -1.566e+00 & 8.521e+00 & -2.830e+00 & 9.755e+00 & -1.996e+00 & 8.629e+00 & \\
3 & 8.552e+00 & 3.716e+00 & -1.242e+00 & 8.845e+00 & -2.512e+00 & 1.007e+01 & -1.394e+00 & 8.545e+00 & \\
4 & 4.930e+00 & 5.456e+00 & 5.887e-01 & 1.068e+01 & -6.879e-01 & 1.190e+01 & 5.693e-02 & 1.060e+01 & \\
5 & 1.029e+01 & 1.208e+01 & 7.423e+00 & 1.751e+01 & 6.165e+00 & 1.875e+01 & 7.167e+00 & 1.778e+01 & \\
6 & 4.178e+01 & 3.702e+01 & 3.351e+01 & 4.366e+01 & 3.252e+01 & 4.510e+01 & 3.298e+01 & 4.470e+01 & \\
7 & 7.783e+01 & 7.820e+01 & 7.447e+01 & 8.551e+01 & 7.439e+01 & 8.697e+01 & 7.480e+01 & 8.636e+01 & \\
8 & 7.844e+01 & 7.709e+01 & 7.236e+01 & 8.332e+01 & 7.230e+01 & 8.488e+01 & 7.263e+01 & 8.453e+01 & \\
9 & 5.618e+01 & 5.556e+01 & 5.085e+01 & 6.130e+01 & 5.027e+01 & 6.286e+01 & 5.121e+01 & 6.225e+01 & \\
10 & 3.780e+01 & 3.691e+01 & 3.207e+01 & 4.217e+01 & 3.113e+01 & 4.371e+01 & 3.188e+01 & 4.271e+01 & \\
11 & 2.325e+01 & 2.402e+01 & 1.912e+01 & 2.921e+01 & 1.798e+01 & 3.056e+01 & 1.900e+01 & 2.969e+01 & \\
12 & 1.937e+01 & 1.565e+01 & 1.070e+01 & 2.079e+01 & 9.467e+00 & 2.205e+01 & 1.030e+01 & 2.094e+01 & \\
13 & 1.317e+01 & 1.034e+01 & 5.354e+00 & 1.544e+01 & 4.086e+00 & 1.667e+01 & 4.917e+00 & 1.561e+01 & \\
14 & 8.096e-01 & 7.036e+00 & 2.019e+00 & 1.211e+01 & 7.399e-01 & 1.333e+01 & 1.946e+00 & 1.230e+01 & \\
15 & 2.381e+00 & 5.030e+00 & -1.277e-02 & 1.007e+01 & -1.292e+00 & 1.129e+01 & -2.094e-01 & 1.021e+01 & \\
16 & 1.747e+00 & 3.898e+00 & -1.170e+00 & 8.917e+00 & -2.445e+00 & 1.014e+01 & -1.033e+00 & 9.236e+00 & \\
17 & 2.904e+00 & 3.435e+00 & -1.674e+00 & 8.413e+00 & -2.940e+00 & 9.645e+00 & -1.793e+00 & 8.531e+00 & \\
18 & 1.409e+00 & 3.700e+00 & -1.480e+00 & 8.607e+00 & -2.732e+00 & 9.853e+00 & -1.505e+00 & 9.062e+00 & \\
19 & 9.303e+00 & 5.394e+00 & -9.628e-03 & 1.008e+01 & -1.227e+00 & 1.136e+01 & 2.618e-01 & 1.056e+01 & \\
20 & 1.114e+01 & 1.184e+01 & 5.597e+00 & 1.568e+01 & 4.578e+00 & 1.716e+01 & 6.865e+00 & 1.730e+01 & \\
21 & 3.388e+01 & 3.622e+01 & 2.765e+01 & 3.862e+01 & 2.740e+01 & 3.998e+01 & 3.055e+01 & 4.150e+01 & \\
22 & 8.328e+01 & 7.771e+01 & 7.197e+01 & 8.306e+01 & 7.193e+01 & 8.452e+01 & 7.272e+01 & 8.509e+01 & \\
23 & 8.053e+01 & 7.745e+01 & 7.579e+01 & 8.654e+01 & 7.556e+01 & 8.814e+01 & 7.390e+01 & 8.533e+01 & \\
24 & 5.640e+01 & 5.599e+01 & 5.433e+01 & 6.441e+01 & 5.317e+01 & 6.576e+01 & 5.213e+01 & 6.339e+01 & \\
25 & 3.562e+01 & 3.722e+01 & 3.459e+01 & 4.468e+01 & 3.316e+01 & 4.575e+01 & 3.307e+01 & 4.380e+01 & \\
26 & 2.464e+01 & 2.423e+01 & 2.078e+01 & 3.087e+01 & 1.930e+01 & 3.188e+01 & 1.956e+01 & 3.038e+01 & \\
27 & 1.785e+01 & 1.578e+01 & 1.176e+01 & 2.185e+01 & 1.029e+01 & 2.288e+01 & 1.096e+01 & 2.153e+01 & \\
28 & 1.065e+01 & 1.042e+01 & 6.021e+00 & 1.611e+01 & 4.582e+00 & 1.717e+01 & 5.093e+00 & 1.595e+01 & \\
29 & 6.680e+00 & 7.087e+00 & 2.427e+00 & 1.251e+01 & 1.029e+00 & 1.361e+01 & 2.178e+00 & 1.252e+01 & \\
30 & 4.822e+00 & 5.060e+00 & 2.266e-01 & 1.031e+01 & -1.136e+00 & 1.145e+01 & 7.373e-02 & 1.025e+01 & \\
\bottomrule
\end{tabular*}
\end{table*}

\subsection{Case III: Isomerization of $\alpha$-Pinene}

As a third case study, we examined the $\alpha$-pinene isomerization model. The isomerization process of $\alpha$-pinene is significant in industry, especially in the production of synthetic fragrances and flavors. These complex biochemical reactions can be effectively modeled using a system of five differential equations with five unknown parameters. The resulting kinetic model has been a classical example in the analysis of multiresponse data \citep{box1973some}. The kinetic equations encapsulate the transformation dynamics of $\alpha$-pinene into its various isomers through a series of reaction steps: 

\begin{equation}\label{eq:ap_eqs}
    \begin{aligned}
        \dot{x}_1&=-(p_1+p_2)x_1, \\
        \dot{x}_2&=p_1x_1,\\
        \dot{x}_3&=p_2x_1-(p_3+p_4)x_3+p_5x_5,\\
        \dot{x}_4&=p_3x_3,\\
        \dot{x}_5&=p_4x_3-p_5x_5.
    \end{aligned}
\end{equation}

In the equations above, each $p_i\in [0,1]$, $i=1,\dots,5$ represents a different rate of reaction, defining the conversion speed from one isomer to another. The initial conditions considered in the generation of the datasets were $x(0)=(100,0,0,0,0)$. Additionally, the values of the parameters used were $p=(5.93e-05, 2.96e-05,2.05e-05,2.75e-04,4.00e-05)$. The dataset generation procedure for this case study mirrored that used for the Logistic model, employing the same noise levels and dataset sizes. Although we generated synthetic datasets to assess the method's behavior, we illustrated this behavior with a real dataset from \cite{box1973some}.

Figure \ref{fig:alpha}
 shows the resulting regions of the isomerization of $\alpha$-Pinene by applying the different algorithms to the 9-point real dataset. The results are once again consistent between both conformal algorithms and closely align with the regions obtained using STAN. In terms of computational cost, the conformal algorithms are notably more efficient, requiring less than a minute to compute the regions, whereas the Bayesian approach takes several minutes.

\begin{center}
    \includegraphics[width=0.7\textwidth]{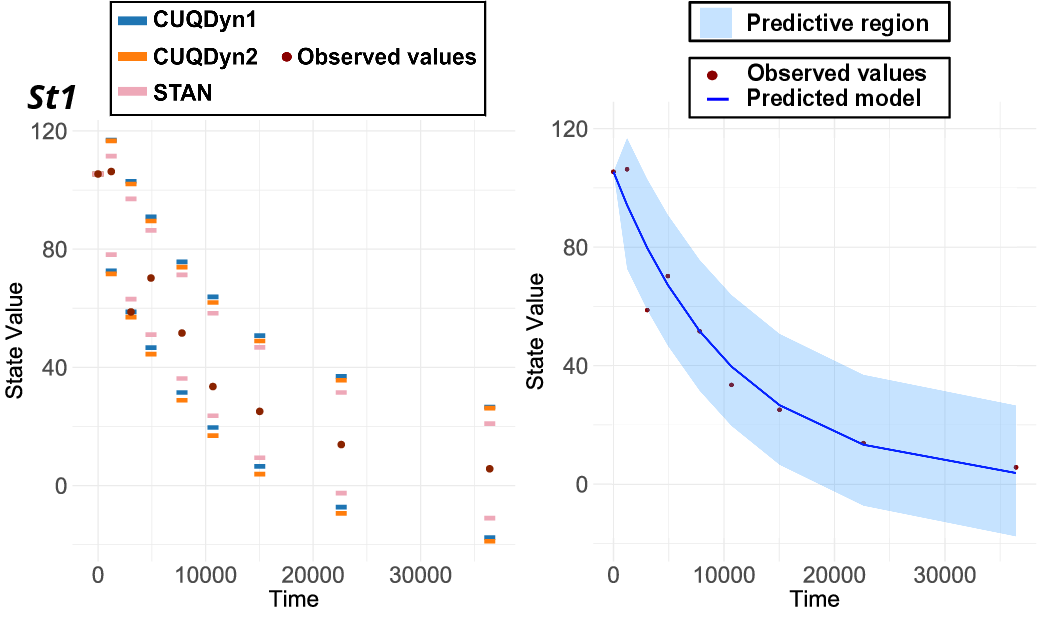}
\end{center}
\begin{center}
    \includegraphics[width=0.7\textwidth]{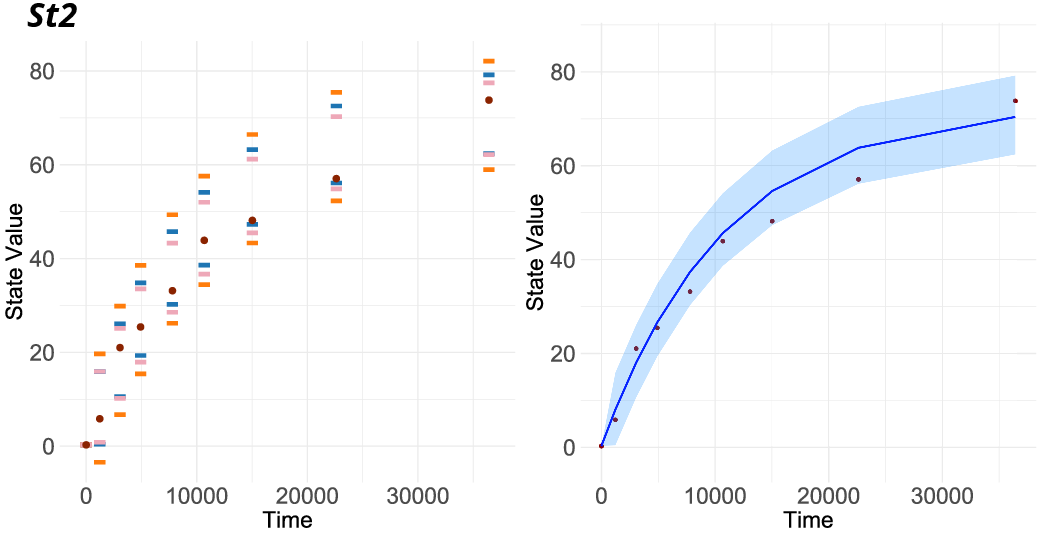}
\end{center}
\begin{center}
    \includegraphics[width=0.7\textwidth]{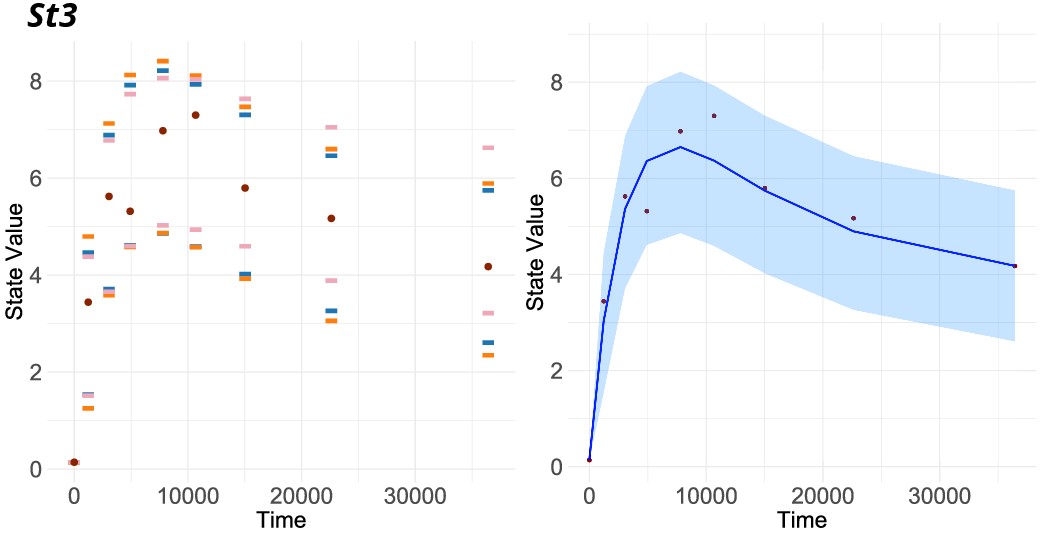}
\end{center}
\begin{center}
    \includegraphics[width=0.7\textwidth]{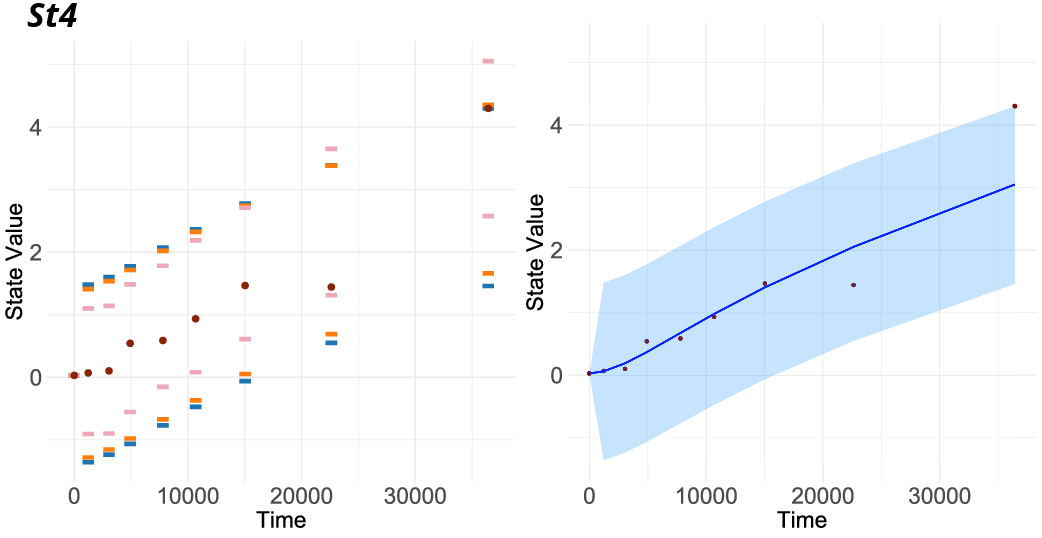}
\end{center}
\begin{figure}[H]
    \centering
    \includegraphics[width=0.7\textwidth]{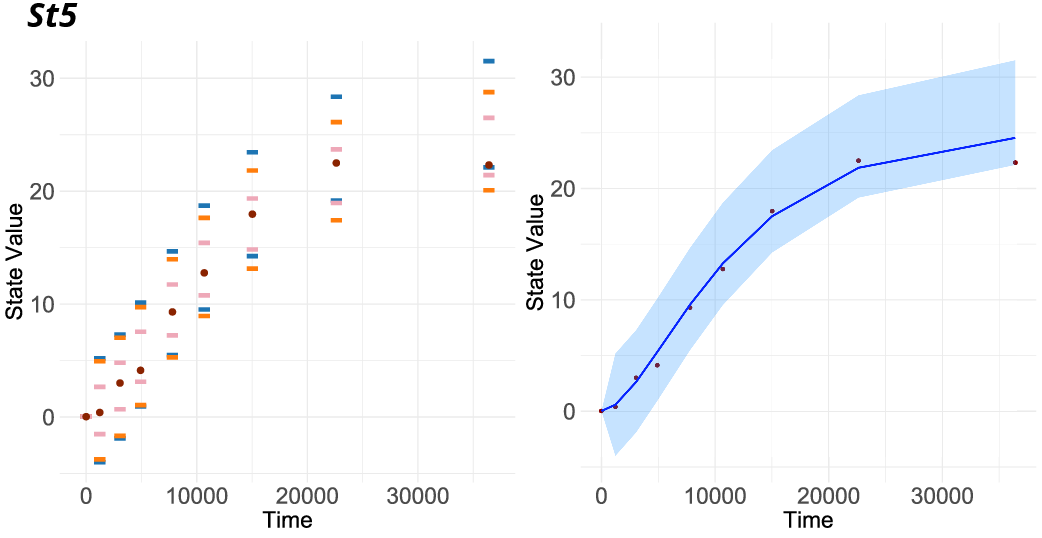}
    \caption{Comparative analysis of the $\alpha$-pinene isomerization model predictive regions. This figure presents the $95\%$ predictive regions obtained from a 9-point real dataset. It showcases the regions for the first two states obtained by using three different methodologies: our two proposed methods (\texttt{CUQDyn1} and \texttt{CUQDyn2}) and a Bayesian approach implemented with STAN.}
    \label{fig:alpha}
\end{figure}

\begin{table*}[!htbp]
\caption{Numerical results corresponding to the comparative analysis of the $\alpha$-pinene isomerization model predictive regions for the first state. This table displays the observed data ($y$) alongside the $95\%$ lower and upper predictive bounds (LPB and UPB, respectively) calculated for each time point ($t$) in a 9-point real dataset. The results are reported for three methodologies: the proposed \texttt{CUQDyn1} and \texttt{CUQDyn2} methods and a Bayesian method implemented with STAN.\label{tab:ap-comp-1}}
\setlength{\tabcolsep}{1.9pt}
\begin{tabular*}{\textwidth}{@{\extracolsep{\fill}}lcccccccc}
\toprule
& & \multicolumn{2}{c}{CUQDyn1} & \multicolumn{2}{c}{CUQDyn2} & \multicolumn{2}{c}{STAN}\\
\cline{3-4}\cline{5-6}\cline{7-8}
$t$ & $y$ & LPB & UPB & LPB & UPB & LPB & UPB \\
\midrule
0 & 1.054e+02 & 1.054e+02 & 1.054e+02 & 1.054e+02 & 1.054e+02 & 1.054e+02 & 1.054e+02 & \\
1230 & 1.063e+02 & 7.266e+01 & 1.169e+02 & 7.162e+01 & 1.167e+02 & 7.814e+01 & 1.115e+02 & \\
3060 & 5.875e+01 & 5.875e+01 & 1.030e+02 & 5.702e+01 & 1.020e+02 & 6.308e+01 & 9.701e+01 & \\
4920 & 7.023e+01 & 4.670e+01 & 9.091e+01 & 4.450e+01 & 8.953e+01 & 5.110e+01 & 8.633e+01 & \\
7800 & 5.161e+01 & 3.149e+01 & 7.570e+01 & 2.888e+01 & 7.391e+01 & 3.624e+01 & 7.128e+01 & \\
10680 & 3.350e+01 & 1.965e+01 & 6.385e+01 & 1.691e+01 & 6.194e+01 & 2.364e+01 & 5.830e+01 & \\
15030 & 2.510e+01 & 6.523e+00 & 5.073e+01 & 3.894e+00 & 4.892e+01 & 9.418e+00 & 4.676e+01 & \\
22620 & 1.388e+01 & -7.284e+00 & 3.693e+01 & -9.388e+00 & 3.564e+01 & -2.531e+00 & 3.148e+01 & \\
36420 & 5.688e+00 & -1.763e+01 & 2.658e+01 & -1.883e+01 & 2.620e+01 & -1.103e+01 & 2.096e+01 & \\
\bottomrule
\end{tabular*}
\end{table*}

\begin{table*}[!htbp]
\caption{Numerical results corresponding to the comparative analysis of the $\alpha$-pinene isomerization model predictive regions for the second state. This table displays the observed data ($y$) alongside the $95\%$ lower and upper predictive bounds (LPB and UPB, respectively) calculated for each time point ($t$) in a 9-point real dataset. The results are reported for three methodologies: the proposed \texttt{CUQDyn1} and \texttt{CUQDyn2} methods and a Bayesian method implemented with STAN.\label{tab:ap-comp-2}}
\setlength{\tabcolsep}{1.9pt}
\begin{tabular*}{\textwidth}{@{\extracolsep{\fill}}lcccccccc}
\toprule
& & \multicolumn{2}{c}{CUQDyn1} & \multicolumn{2}{c}{CUQDyn2} & \multicolumn{2}{c}{STAN}\\
\cline{3-4}\cline{5-6}\cline{7-8}
$t$ & $y$ & LPB & UPB & LPB & UPB & LPB & UPB \\
\midrule
0 & 2.769e-01 & 2.769e-01 & 2.769e-01 & 2.769e-01 & 2.769e-01 & 2.769e-01 & 2.769e-01 & \\
1230 & 5.842e+00 & 4.078e-01 & 1.589e+01 & -3.424e+00 & 1.970e+01 & 8.377e-01 & 1.593e+01 & \\
3060 & 2.101e+01 & 1.059e+01 & 2.608e+01 & 6.730e+00 & 2.985e+01 & 1.020e+01 & 2.511e+01 & \\
4920 & 2.542e+01 & 1.934e+01 & 3.482e+01 & 1.543e+01 & 3.855e+01 & 1.792e+01 & 3.353e+01 & \\
7800 & 3.314e+01 & 3.024e+01 & 4.573e+01 & 2.621e+01 & 4.933e+01 & 2.855e+01 & 4.330e+01 & \\
10680 & 4.389e+01 & 3.862e+01 & 5.410e+01 & 3.444e+01 & 5.756e+01 & 3.669e+01 & 5.201e+01 & \\
15030 & 4.814e+01 & 4.729e+01 & 6.321e+01 & 4.333e+01 & 6.645e+01 & 4.547e+01 & 6.119e+01 & \\
22620 & 5.705e+01 & 5.611e+01 & 7.253e+01 & 5.231e+01 & 7.543e+01 & 5.487e+01 & 7.025e+01 & \\
36420 & 7.378e+01 & 6.237e+01 & 7.918e+01 & 5.897e+01 & 8.209e+01 & 6.217e+01 & 7.745e+01 & \\
\bottomrule
\end{tabular*}
\end{table*}

\begin{table*}[!htbp]
\caption{Numerical results corresponding to the comparative analysis of the $\alpha$-pinene isomerization model predictive regions for the third state. This table displays the observed data ($y$) alongside the $95\%$ lower and upper predictive bounds (LPB and UPB, respectively) calculated for each time point ($t$) in a 9-point real dataset. The results are reported for three methodologies: the proposed \texttt{CUQDyn1} and \texttt{CUQDyn2} methods and a Bayesian method implemented with STAN.\label{tab:ap-comp-3}}
\setlength{\tabcolsep}{1.9pt}
\begin{tabular*}{\textwidth}{@{\extracolsep{\fill}}lcccccccc}
\toprule
& & \multicolumn{2}{c}{CUQDyn1} & \multicolumn{2}{c}{CUQDyn2} & \multicolumn{2}{c}{STAN}\\
\cline{3-4}\cline{5-6}\cline{7-8}
$t$ & $y$ & LPB & UPB & LPB & UPB & LPB & UPB \\
\midrule
0 & 1.409e-01 & 1.409e-01 & 1.409e-01 & 1.409e-01 & 1.409e-01 & 1.409e-01 & 1.409e-01 & \\
1230 & 3.441e+00 & 1.535e+00 & 4.463e+00 & 1.253e+00 & 4.796e+00 & 1.515e+00 & 4.380e+00 & \\
3060 & 5.622e+00 & 3.709e+00 & 6.887e+00 & 3.584e+00 & 7.127e+00 & 3.656e+00 & 6.776e+00 & \\
4920 & 5.316e+00 & 4.615e+00 & 7.918e+00 & 4.583e+00 & 8.125e+00 & 4.601e+00 & 7.730e+00 & \\
7800 & 6.977e+00 & 4.858e+00 & 8.217e+00 & 4.869e+00 & 8.411e+00 & 5.025e+00 & 8.063e+00 & \\
10680 & 7.298e+00 & 4.590e+00 & 7.931e+00 & 4.574e+00 & 8.117e+00 & 4.936e+00 & 8.036e+00 & \\
15030 & 5.795e+00 & 4.021e+00 & 7.303e+00 & 3.928e+00 & 7.470e+00 & 4.595e+00 & 7.635e+00 & \\
22620 & 5.169e+00 & 3.261e+00 & 6.461e+00 & 3.055e+00 & 6.598e+00 & 3.885e+00 & 7.048e+00 & \\
36420 & 4.175e+00 & 2.608e+00 & 5.748e+00 & 2.346e+00 & 5.889e+00 & 3.215e+00 & 6.627e+00 & \\
\bottomrule
\end{tabular*}
\end{table*}

\begin{table*}[!htbp]
\caption{Numerical results corresponding to the comparative analysis of the $\alpha$-pinene isomerization model predictive regions for the fourth state. This table displays the observed data ($y$) alongside the $95\%$ lower and upper predictive bounds (LPB and UPB, respectively) calculated for each time point ($t$) in a 9-point real dataset. The results are reported for three methodologies: the proposed \texttt{CUQDyn1} and \texttt{CUQDyn2} methods and a Bayesian method implemented with STAN.\label{tab:ap-comp-4}}
\setlength{\tabcolsep}{1.9pt}
\begin{tabular*}{\textwidth}{@{\extracolsep{\fill}}lcccccccc}
\toprule
& & \multicolumn{2}{c}{CUQDyn1} & \multicolumn{2}{c}{CUQDyn2} & \multicolumn{2}{c}{STAN}\\
\cline{3-4}\cline{5-6}\cline{7-8}
$t$ & $y$ & LPB & UPB & LPB & UPB & LPB & UPB \\
\midrule
0 & 3.034e-02 & 3.034e-02 & 3.034e-02 & 3.034e-02 & 3.034e-02 & 3.034e-02 & 3.034e-02 & \\
1230 & 6.992e-02 & -1.359e+00 & 1.482e+00 & -1.283e+00 & 1.411e+00 & -9.063e-01 & 1.103e+00 & \\
3060 & 1.027e-01 & -1.237e+00 & 1.604e+00 & -1.156e+00 & 1.539e+00 & -8.984e-01 & 1.143e+00 & \\
4920 & 5.430e-01 & -1.065e+00 & 1.776e+00 & -9.783e-01 & 1.716e+00 & -5.551e-01 & 1.485e+00 & \\
7800 & 5.887e-01 & -7.679e-01 & 2.073e+00 & -6.727e-01 & 2.022e+00 & -1.521e-01 & 1.784e+00 & \\
10680 & 9.356e-01 & -4.725e-01 & 2.368e+00 & -3.696e-01 & 2.325e+00 & 8.084e-02 & 2.190e+00 & \\
15030 & 1.468e+00 & -6.209e-02 & 2.779e+00 & 5.318e-02 & 2.747e+00 & 6.100e-01 & 2.711e+00 & \\
22620 & 1.443e+00 & 5.493e-01 & 3.390e+00 & 6.911e-01 & 3.385e+00 & 1.312e+00 & 3.652e+00 & \\
36420 & 4.301e+00 & 1.460e+00 & 4.301e+00 & 1.662e+00 & 4.356e+00 & 2.578e+00 & 5.056e+00 & \\
\bottomrule
\end{tabular*}
\end{table*}

\begin{table*}[!htbp]
\caption{Numerical results corresponding to the comparative analysis of the $\alpha$-pinene isomerization model predictive regions for the fifth state. This table displays the observed data ($y$) alongside the $95\%$ lower and upper predictive bounds (LPB and UPB, respectively) calculated for each time point ($t$) in a 9-point real dataset. The results are reported for three methodologies: the proposed \texttt{CUQDyn1} and \texttt{CUQDyn2} methods and a Bayesian method implemented with STAN.\label{tab:ap-comp-5}}
\setlength{\tabcolsep}{1.9pt}
\begin{tabular*}{\textwidth}{@{\extracolsep{\fill}}lcccccccc}
\toprule
& & \multicolumn{2}{c}{CUQDyn1} & \multicolumn{2}{c}{CUQDyn2} & \multicolumn{2}{c}{STAN}\\
\cline{3-4}\cline{5-6}\cline{7-8}
$t$ & $y$ & LPB & UPB & LPB & UPB & LPB & UPB \\
\midrule
0 & 3.252e-02 & 3.252e-02 & 3.252e-02 & 3.252e-02 & 3.252e-02 & 3.252e-02 & 3.252e-02 & \\
1230 & 4.046e-01 & -4.002e+00 & 5.192e+00 & -3.748e+00 & 4.940e+00 & -1.514e+00 & 2.666e+00 & \\
3060 & 3.011e+00 & -1.894e+00 & 7.300e+00 & -1.668e+00 & 7.020e+00 & 6.917e-01 & 4.812e+00 & \\
4920 & 4.136e+00 & 9.374e-01 & 1.013e+01 & 1.051e+00 & 9.739e+00 & 3.127e+00 & 7.557e+00 & \\
7800 & 9.308e+00 & 5.485e+00 & 1.468e+01 & 5.285e+00 & 1.397e+01 & 7.238e+00 & 1.173e+01 & \\
10680 & 1.277e+01 & 9.528e+00 & 1.872e+01 & 8.952e+00 & 1.764e+01 & 1.078e+01 & 1.542e+01 & \\
15030 & 1.797e+01 & 1.426e+01 & 2.345e+01 & 1.315e+01 & 2.184e+01 & 1.484e+01 & 1.935e+01 & \\
22620 & 2.250e+01 & 1.917e+01 & 2.837e+01 & 1.743e+01 & 2.612e+01 & 1.895e+01 & 2.370e+01 & \\
36420 & 2.232e+01 & 2.212e+01 & 3.152e+01 & 2.008e+01 & 2.877e+01 & 2.142e+01 & 2.650e+01 & \\
\bottomrule
\end{tabular*}
\end{table*}

\begin{sidewaystable*}[!htbp]
\caption{Parameter estimation comparison for the first three case studies. Since the parameter estimation process is identical for both algorithms proposed in this paper, for simplicity, we will refer to the parameters obtained with them as CUQDyn. For the case study on the isomerization of $\alpha$-pinene, a real dataset was considered, and thus the nominal parameters for this problem are unknown.\label{tab3}}
\tabcolsep=0pt
\begin{tabular*}{\textwidth}{@{\extracolsep{\fill}}lccccccccccc@{\extracolsep{\fill}}}
\toprule%
& \multicolumn{2}{@{}c@{}}{Logistic} & \multicolumn{4}{@{}c@{}}{Lotka-Volterra} & \multicolumn{5}{@{}c@{}}{$\alpha$-Pinene} \\
\cline{2-3}\cline{4-7}\cline{8-12}%
 & r & K & $\alpha$ & $\beta$ & $\gamma$ & $\delta$ & $p_1$ & $p_2$ & $p_3$ & $p_4$ & $p_5$ \\
\midrule
True  & 1.000e-01 & 1.000e+02 & 5.000e-01 & 2.000e-02 & 5.000e-01 & 2.000e-02 & NA & NA & NA & NA & NA \\
STAN  & 1.152e-01 & 1.021e+02 & 4.992e-01 & 2.004e-02 & 5.000e-01 & 2.001e-02 & 5.998e-05 & 2.797e-05 & 1.860e-05 & 2.826e-04 & 5.366e-05\\
CUQDyn  & 1.112e-01 & 1.027e+02 & 4.984e-01 & 2.001e-02 & 5.001e-01 & 2.000e-02 & 6.307e-05 & 2.841e-05 & 1.609e-05 & 2.729e-04 & 4.365e-05\\
\bottomrule
\end{tabular*}
\end{sidewaystable*}

\newpage
\clearpage
\subsection{Case IV: NFKB signaling pathway}

The Nuclear Factor Kappa-light-chain-enhancer of activated B cells (NFKB) signaling pathway plays a key role in the regulation of immune response, inflammation and cell survival. This pathway is activated in response to various stimuli, including cytokines, stress and microbial infections, leading yo the transcription of target genes involved in immune and inflammatory responses.
Here we consider the dynamics of this pathway as described by a system of differential equations \citep{lipniacki-paszek-brasier-luxon-kimmel:2004}:

\begin{equation}\label{eq:nfkb_eqs}
    \begin{aligned}
        \dot{IKKn} &= kprod - kdeg \cdot IKKn\\
        &- Tr \cdot k1 \cdot IKKn, \\
        \dot{IKKa} &= Tr \cdot k1 \cdot IKKn - k3 \cdot IKKa \\
        &- Tr \cdot k2 \cdot IKKa \cdot A20 - kdeg \cdot IKKa\\
        &- a2 \cdot IKKa \cdot IkBa + t1 \cdot IKKaIkBa\\
        &- a3 \cdot IKKa \cdot IkBaNFkB\\
        &+ t2 \cdot IKKaIkBaNFkB, \\
        \dot{IKKi} &= k3 \cdot IKKa + Tr \cdot k2 \cdot IKKa \cdot A20\\
        &- kdeg \cdot IKKi, \\
        \dot{IKKaIkBa} &= a2 \cdot IKKa \cdot IkBa - t1 \cdot IKKaIkBa, \\
        \dot{IKKaIkBaNFkB} &= a3 \cdot IKKa \cdot IkBaNFkB\\
        &- t2 \cdot IKKaIkBaNFkB, \\
        \dot{NFkB} &= c6a \cdot IkBaNFkB - a1 \cdot NFkB \cdot IkBa\\
        &+ t2 \cdot IKKaIkBaNFkB - i1 \cdot NFkB, \\
        \dot{NFkBn} &= i1 \cdot kv \cdot NFkB - a1 \cdot IkBan \cdot NFkBn, \\
        \dot{A20} &= c4 \cdot A20t - c5 \cdot A20, \\
        \dot{A20t} &= c2 + c1 \cdot NFkBn - c3 \cdot A20t, \\
        \dot{IkBa} &= -a2 \cdot IKKa \cdot IkBa\\
        &- a1 \cdot IkBa \cdot NFkB\\
        &+ c4a \cdot IkBat - c5a \cdot IkBa - i1a \cdot IkBa\\
        &+ e1a \cdot IkBan, \\
        \dot{IkBan} &= -a1 \cdot IkBan \cdot NFkBn + i1a \cdot kv \cdot IkBa\\
        &- e1a \cdot kv \cdot IkBan, \\
        \dot{IkBat} &= c2a + c1a \cdot NFkBn - c3a \cdot IkBat, \\
        \dot{IkBaNFkB} &= a1 \cdot IkBa \cdot NFkB - c6a \cdot IkBaNFkB\\
        & - a3 \cdot IKKa \cdot IkBaNFkB\\
        &+ e2a \cdot IkBanNFkBn, \\
        \dot{IkBanNFkBn} &= a1 \cdot IkBan \cdot NFkBn\\
        &- e2a \cdot kv \cdot IkBanNFkBn, \\
        \dot{cgent} &= c2c + c1c \cdot NFkBn - c3c \cdot cgent.
    \end{aligned}
\end{equation}

In agreement with the scenario considered by \cite{}, we assume that the available measurements are determined by the observation function \( g: \mathbb{R}^{15} \to \mathbb{R}^{6} \), which is defined as follows:

\begin{equation}\label{eq:obs_func}
    \begin{aligned}
       g(\cdot) =& (NFkBn(\cdot), IkBa(\cdot) + IkBaNFkB(\cdot), A20t(\cdot),\\
        &IKKn(\cdot) + IKKa(\cdot) + IKKi(\cdot), IKKa(\cdot), IkBat(\cdot)).
    \end{aligned}
\end{equation}

Out of the system of 15 equations, only 6 observables, defined by the function $g$, are available. The parameter values used in the generation of the datasets are as follows:

\begin{align*}
\text{a1} &= 5e-01, & \text{a2} &= 2e-01, & \text{t1} &= 1e-01, \\
\text{a3} &= 1e+00, & \text{t2} &= 1e-01,& \text{c1a} &= 5e-07, \\
\text{c2a} &= 0e+00, & \text{c3a} &= 4e-04, & \text{c4a} &= 5e-01, \\
\text{c5a} &= 1e-04, & \text{c6a} &= 2e-05, &\text{c1} &= 5e-07,\\
\text{c2} &= 0e+00, & \text{c3} &= 4e-04, & \text{c4} &= 5e-01, \\
\text{c5} &= 3e-04, & \text{k1} &= 2.5e-03, & \text{k2} &= 1e-01, \\
\text{k3} &= 1.5e-03, & \text{kprod} &= 2.5e-05, &\text{kdeg} &= 1.25e-04, \\
\text{kv} &= 5e+00, & \text{i1} &= 2.5e-03, & \text{e2a} &= 1e-02, \\
\text{i1a} &= 1e-03, & \text{e1a} &= 5e-04, &\text{c1c} &= 5e-07,\\
\text{c2c} &= 0e+00, & \text{c3c} &= 4e-04.
\end{align*}

\begin{figure}[!hhh]
    \centering
    \includegraphics[width=0.5\textwidth]{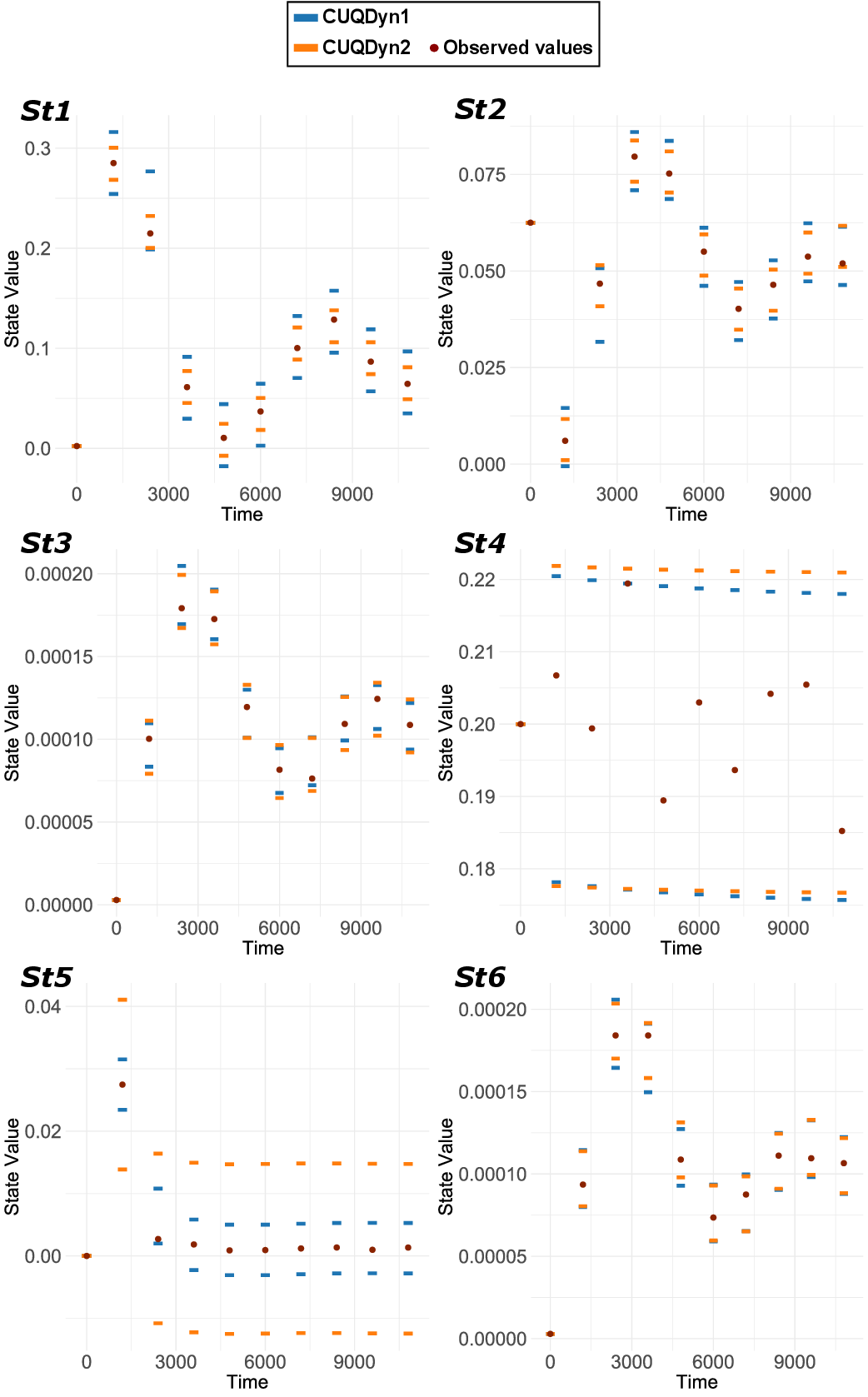}
    \caption{Comparative analysis of the NFkb signaling pathway model predictive regions. This figure presents the $95\%$ predictive regions obtained from a 13-point synthetic dataset. It showcases the regions for the six observables obtained by using our two proposed methods: \texttt{CUQDyn1} and \texttt{CUQDyn2}.\\
    \\}
    \label{fig:nfkb_plot}
\end{figure}

It should be noted that in this problem, which involves 29 unknown parameters and 15 state variables, we only have access to 6 observable outputs. This discrepancy between the number of parameters and the available observables presents a challenge in the context of parameter identifiability, and is very commmon in systems biology applications. Identifiability refers to the ability to uniquely determine the model parameters based on the available data. When a system lacks identifiability, inferring unique parameter values from observable data becomes challenging, if not impossible. However, as mentioned in the introduction, by characterizing the impact of this lack of identifiability using appropriate uncertainty quantification (UQ) methods, it might still be possible to make useful predictions.

Figure \ref{fig:nfkb_plot}
    shows the results of applying our two methods to a 13-point synthetic dataset. Both methods based on conformal inference yielded results that are in close agreement with each other. However, in this case, we were not able to obtain adequate predictive regions using STAN, even after many hours of computation, probably due to the partial lack of identifiability. Remarkably, our \texttt{CUQDyn1} and \texttt{CUQDyn2} algorithms can compute the regions in just a few minutes using a standard PC.

The dataset generation process was the same that the one used in the previous case studies.

\end{appendix}

\end{document}